\documentclass[oneside, a4paper, 12pt, bibliography=totoc]{scrartcl}
\usepackage[a4paper]{anysize}
\usepackage{marginnote}
\usepackage{times}
\usepackage[hyphens]{url}
\usepackage{hyperref}
\usepackage[round,colon]{natbib}
\usepackage[utf8]{inputenc}
\usepackage[fleqn]{amsmath}
\usepackage{amsfonts}
\usepackage{amssymb}
\usepackage{mleftright}
\usepackage{amsthm}
\usepackage{thmtools}
\usepackage{mathtools}
\usepackage{graphicx}
\usepackage{calc}
\declaretheorem[name=Lemma]{lemma}
\begin{document}
\setlength{\mathindent}{0mm}
\newlength{\symaeqmargin}
\setlength{\symaeqmargin}{2cm}
\newlength{\symaeqwidth}
\setlength{\symaeqwidth}{\textwidth - \symaeqmargin}
\allowdisplaybreaks

\newcommand{\mb}[1]{\mathbf{#1}}
\newcommand{\bs}[1]{\boldsymbol{#1}}
\newcommand{\opnl}[1]{\operatorname{#1}\nolimits}
\setlength\nulldelimiterspace{0pt}
\title{Derivation of Learning Rules for Coupled Principal Component
Analysis in a Lagrange-Newton Framework}

\author{Ralf M\"oller\\Computer Engineering, Faculty of
Technology\\Bielefeld University\\\url{www.ti.uni-bielefeld.de}}

\date{version of \today}

\maketitle

\begin{abstract}

\noindent We describe a Lagrange-Newton framework for the derivation of
learning rules with desirable convergence properties and apply it to the
case of principal component analysis (PCA). In this framework, a Newton
descent is applied to an extended variable vector which also includes
Lagrange multipliers introduced with constraints. The Newton descent
guarantees equal convergence speed from all directions, but is also
required to produce stable fixed points in the system with the extended
state vector. The framework produces ``coupled'' PCA learning rules
which simultaneously estimate an eigenvector and the corresponding
eigenvalue in cross-coupled differential equations. We demonstrate the
feasibility of this approach for two PCA learning rules, one for the
estimation of the principal, the other for the estimate of an arbitrary
eigenvector-eigenvalue pair (eigenpair). \end{abstract}

\newpage

\tableofcontents

\newpage

\section{Introduction}

%
There is a plethora of neural network algorithms for principal component
analysis (PCA) \citep[recent textbook:][]{nn_Kong17}. Particularly the
online forms of these algorithms are attractive since they avoid the
construction of the potentially very large covariance matrix. Instead,
eigenvectors and eigenvalues are directly estimated from a data stream,
reducing computational effort and memory demand.

Our main contribution to this field are ``coupled'' learning rules for
PCA \citep[][]{own_Moeller04a}. These rules simultaneously estimate
eigenvectors and eigenvalues in a coupled system of equations, i.e. the
eigenvalue estimates affect the eigenvector update and vice versa.
Coupled learning rules mitigate the speed-stability problem which
affects other learning rules and make the rules independent on the range
of eigenvalues. In our original publication, the rules were derived by
applying Newton's method to an information criterion related to the
normalized Mahalanobis distance. Newton's method ensures approximately
equal convergence speed from all directions, at least in the vicinity of
the fixed points. Moreover, by choosing the Hessian matrix for a desired
fixed point, this fixed point can be turned into an attractor.

The weak point of this approach is the information criterion
\citep[][]{own_Moeller20}. While there is considerable freedom in
designing the criterion --- it just needs to exhibit fixed points in the
eigenvector and eigenvalues, but these don't have to be attractors ---
there is no systematic way to design such a criterion. Moreover, the
criterion has limited explanatory value, since it is not related to an
optimization problem. We therefore suggested to derive coupled learning
rules from a Newton zero-finding framework \citep[][]{own_Moeller20}. In
this approach, an equation derived from the optimum of an objective
function is combined with a constraint on the eigenvector estimates
("weight vectors"). This joint vector is set to zero, and the equation
is solved by applying Newton's method as a zero finder. We managed to
derive PCA and SVD (singular value decomposition) learning rules from
this framework, for both Euclidean and constant-sum constraints of the
eigenvector estimates. However, the framework is not entirely
satisfactory since the approach may fail if the constraints do not
intersect the solution derived from the unconstrained optimum.

Where therefore already indicated in our previous work that switching to
a Lagrange-Newton framework may be a more universal approach
\citep[][]{own_Moeller20}. The present paper is devoted to exploring
this alternative. In the Lagrange-Newton framework, an objective
function is linked to constraints through Lagrange multipliers, and
Newton's method is applied to the resulting extended objective function
by including both the original variables (eigenvector estimates) and the
Lagrange multipliers in the variable vector. It turns out that the
Lagrange multipliers coincide with eigenvalues in the fixed points. Here
Newton's method serves an additional purpose: Solutions of the extended
objective functions are saddle points if the Lagrange multipliers are
included \citep[][]{nn_Kalman09,nn_Baker92,nn_Walsh75}; see also section
\ref{sec_bordered_hessian}. Only through the application of Newton's
method can saddle points be turned into attractors.

We restrict our present investigation to PCA methods and e.g. exclude
SVD methods \citep[coupled methods:][]{own_Kaiser10,own_Moeller20} for
the time being. Moreover, we only study {\em single-component} learning
rules which estimate a single eigenvector / eigenvalue pair (in short
``eigenpair''). We describe two approaches how these single-component
rules can be combined for the estimation of {\em multiple} eigenpairs.
The first approach is the classical deflation operation; the second
approach results immediately from the estimation of an arbitrary (rather
than just the principal) eigenpair, as described below.

In this paper, we only derive learning rules in {\em averaged form}
operating on the covariance matrix. Online rules can always be obtained
by replacing the covariance matrix by the outer product of the current
data vector with itself and introducing a cooling scheme for the
learning rates \citep[see e.g.][]{own_Moeller04a}.

We derive coupled PCA learning rules for two cases: estimation of the
{\em principal} eigenpair and estimation of an {\em arbitrary}
eigenpair. The first case leads to the learning rules known from our
previous work \citep[][]{own_Moeller04a}; these can be integrated into a
deflation scheme to derive multiple eigenpairs. For the second case we
derived a multi-component learning rule where each state relies on the
eigenpair estimates of all {\em previous} stages. The terms relating to
deflation in the first case or to the multi-stage scheme in the second
case differ from each other. For both cases, we provide a stability
analysis.

After introducing the notation in section \ref{sec_notation}, we outline
the Lagrange-Newton approach in section \ref{sec_approach}. The specific
criterion and constraint for the PCA case is introduced in section
\ref{sec_criterion_constraint_pca}. Then we derive a general Hessian
covering both cases studied later in section \ref{sec_hessian}. In
section \ref{sec_stability_exact} we explore the stability of fixed
points under the assumption that the Hessian is known {\em exactly}. The
case of estimating the principal eigenpair is covered in section
\ref{sec_principal}, the case of estimating an arbitrary eigenpair in
section \ref{sec_arbitrary}. Conclusions and future work are presented
in section \ref{sec_conclusion}.

\section{Notation}\label{sec_notation}

%
\textbf{Dimensions:}
\begin{description}
\setlength{\itemsep}{0pt}
\item[] $n$: data dimension, $4 \leq n$
\item[] $m$: number of eigenpair estimates, $4 \leq m \leq n$
\item[] $p$: eigenpair index, typically index related to desired fixed point, $3 \leq p \leq m - 1$
\item[] $q$: eigenpair index, typically index related to some fixed point, $1 \leq q \leq m$
\end{description}
Note that the range restrictions of the dimensions are necessary to
avoid invalid matrix dimensions in some derivation steps.

{\textbf{Matrices:}}
\begin{description}
\setlength{\itemsep}{0pt}
\item[] $\mb{C}$: covariance matrix, symmetric, $n \times n$
\item[] $\check{\mb{C}}_{p-1}$: $p-1$-fold deflated covariance matrix, symmetric, $n \times n$
\item[] $\mb{V}$: true eigenvectors of $\mb{C}$, orthogonal, $n \times n$
\item[] $\bs{\Lambda}$: true eigenvalues of $\mb{C}$, distinct, sorted in descending order, diagonal, $n \times n$
\item[] $\bs{\Lambda}_{p-1}$: upper left block of $\bs{\Lambda}$, diagonal, $(p - 1) \times (p - 1)$
\item[] $\bs{\Lambda}_{n-p}$: lower right block of $\bs{\Lambda}$, diagonal, $(n - p) \times (n - p)$
\item[] $\mb{w}$: weight vector, eigenvector estimate, arbitrary, $n \times 1$
\item[] $l$: Lagrange multiplier, eigenvalue estimate, scalar, $1 \times 1$
\item[] $\mb{W}$: weight vector matrix, arbitrary, $n \times m$
\item[] $\mb{w}_{i}$: weight vector, arbitrary, $n \times 1$, column vector of $\mb{W}$
\item[] $\mb{L}$: eigenvalue estimates, diagonal, $m \times m$
\item[] $l_{i}$: eigenvalue estimate, scalar, $1 \times 1$, diagonal element of $\mb{L}$
\item[] $J$: Lagrange criterion, scalar, $1 \times 1$
\item[] $\mb{T}$: transformation matrix, orthogonal, $(n + 1) \times (n + 1)$
\item[] $\mb{P}_{1,n+1}$: elementary permutation matrix between $1$ and $n+1$, permutation, $(n + 1) \times (n + 1)$
\item[] $\mb{P}_{p,n+1}$: elementary permutation matrix between $p$ and $n+1$, permutation, $(n + 1) \times (n + 1)$
\item[] $\mb{H}$: Hessian matrix, square, $(n + 1) \times (n + 1)$
\item[] ${\mb{H}^*}$: transformed Hessian matrix, square, $(n + 1) \times (n + 1)$
\item[] $\mb{E}$: identity matrix, unit, $n \times n$
\item[] $\mb{e}_{i}$: unit vector with element 1 at position $i$, arbitrary, $n \times 1$, column vector of $\mb{E}$
\end{description}
\textbf{Operators:}
\begin{description}
\setlength{\itemsep}{0pt}
\item[] $\opnl{tr}\{\mb{A}\}$: trace of square matrix $\mb{A}$
\item[] $\opnl{diag}\{\mb{A}\}$: diagonal matrix with diagonal elements from square matrix $\mb{A}$
\item[] $\opnl{sut}\{\mb{A}\}$: strict upper triangular matrix with elements from square matrix $\mb{A}$
\item[] $\opnl{det}\{\mb{A}\}$: determinant of square matrix $\mb{A}$
\end{description}

\section{Lagrange-Newton Approach}\label{sec_approach}

%
While in the original paper \citep[][]{own_Moeller04a} the derivation
started from an ``information criterion'', we now use a
Lagrange-multiplier approach. We will first present the general
approach, but already refer to the specific PCA problem.

Starting point is an \textbf{unconstrained criterion} which depends on
the selected variables. In the PCA case, the criterion is the
expectation of the projected variance; the variables are the eigenvector
estimates (weight vectors); see section
\ref{sec_criterion_constraint_pca}.

We define an \textbf{equality constraint} on the variables. Here we
assume that the constraint is fulfilled if it becomes zero. In the PCA
case, this is the case if the weight vector has unit Euclidean length;
see section \ref{sec_criterion_constraint_pca}.

Criterion and equality constraint are combined in a modified
\textbf{Lagrange criterion}. This introduces Lagrange multiplier
variables for each component of the constraint; see section
\ref{sec_criterion_constraint_pca}. For single-unit PCA, we have a
single Lagrange multiplier; see section \ref{sec_hessian}. From this
point on, we operate on an extended variable vector which includes both
the original variables and the Lagrange multipliers.

The \textbf{set of fixed points} is determined from the zero points of
the first-order derivatives of the Lagrange criterion with respect to
the extended variable vector. In the PCA case, the original variables
have solutions in the eigenvectors, the Lagrange multiplier in the
corresponding eigenvalues of the covariance matrix; see section
\ref{sec_hessian}.

In the Lagrange-Newton framework, we derive the update equation from a
\textbf{Newton descent} on the extended variable vector. As mentioned
above, Newton's method turns the selected saddle point of the Lagrange
function into an attractor. In addition, the convergence speed is
approximately $-1$ from all directions; see section
\ref{sec_stability_exact}.

For the Newton descent, we \textbf{select the desired fixed point}. This
is accomplished \textbf{by computing the Hessian for this fixed point}.
In this work, we explore PCA solutions for the {\em principal} eigenpair
(section \ref{sec_principal}) and for {\em arbitrary} eigenpairs
(section \ref{sec_arbitrary}).

A crucial step in the computation of the Hessian for the desired fixed
point are \textbf{approximations}. Currently, this is the least formal
step of the method. We can identity three aspects:

\begin{itemize}

\item Since we need to invert the Hessian, we have to transform it to a
\textbf{simple form}. In the PCA case, we apply an orthogonal similarity
transformation which diagonalizes a sub-matrix; see section
\ref{sec_hessian}. For this step, \textbf{substitutions} are required:
The extended variable vector is temporarily substituted by the
corresponding fixed-point vector since this enables some
simplifications. This rests on the assumption that the Newton descent
has moved the estimate of the extended variable vector to the vicinity
of the desired fixed point. After the inversion of the Hessian, all
fixed-point quantities are substituted back to the variable quantities
--- the fixed-point quantities are unknown, whereas the variable
quantities are determined in the Newton descent.

\item For further simplification, we \textbf{apply knowledge on the
fixed points} of the extended variable vector. In the PCA case, we
assume an ordered set of eigenvalues where preceding eigenvalues are
much larger than following eigenvalues. This leads to some
\textbf{approximations} which further simplify the Hessian. We noticed
that this step is critical in the PCA case: Approximations are required
to eliminate unknown estimates (following eigenpairs), but may lead to
undesired fixed points if also applied to known estimates (preceding
eigenpairs). We therefore recommend to only use approximations to
eliminate unknown quantities.

\item We have to \textbf{eliminate unknown fixed-point variables}. To
determine the {\em principal} eigenpair in the PCA case (section
\ref{sec_principal}), we assume that all non-principal eigenpairs are
unknown. To determine an {\em arbitrary} eigenpair, we assume that there
are learning rules which estimate all previous eigenpairs, but that all
following eigenpairs are unknown; all fixed-point quantities relating to
the following eigenpairs need to be eliminated. It may not always be
obvious how to achieve this, since the matrix used in the orthogonal
similarity transformation also contains unknown quantities; the
transformed Hessian needs to be brought into a form where the unknown
quantities disappear in the inverse transformation.

\end{itemize}

Finally, we \textbf{invert the approximated Hessian} and perform the
back-substitution mentioned above. We then combine the inverted
approximated Hessian with the gradient to obtain the Newton descent in
the form of an \textbf{ordinary differential equation} over the extended
variable vector. In the context of neural networks, this equation can be
called a ``learning rule''.

Note that usually the Newton descent does not change the set of fixed
points as expressed by the gradient. If the inverse Hessian
$\mb{H}^{-1}$ is defined, the fixed-point equation of the Newton descent
could be multiplied from the left by $\mb{H}$ which is obviously
non-singular, leading to the original fixed-point equation. However, for
the case of arbitrary eigenpairs explored in section
\ref{sec_arbitrary}, the Hessian is singular and thus the inverse
Hessian not defined in some of the original fixed points. This also
introduces an additional set of fixed points.

\section{Criterion and Constraint}\label{sec_criterion_constraint_pca}

%
As the objective function for PCA we use the the variance of the
projection onto the weight vector $\mb{w}$; this objective function is
maximized. This leads to Rayleigh quotient \citep[see e.g.][for the
derivation]{own_Moeller20}:
\begin{align}
\label{eq_rayleigh}
J
&=
\frac{1}{2}\, \frac{\mb{w}^T \mb{C} \mb{w} }{\mb{w}^T \mb{w} }\,.
\end{align}
Instead of normalizing the weight vectors within the Rayleigh quotient,
we can use the numerator of the Rayleigh quotient as objective function
and force the weight vector to L2 unit length by an equality
constraint.\footnote{We don't address the problem of using different
constraints, e.g. the constant sum of elements
\citep[][]{own_Moeller20}. If a constraint other than the L2 unit length
is used, the modified objective function should combine it with the
Rayleigh quotient \eqref{eq_rayleigh} rather than just its numerator.}
The constraint is integrated into the modified objective function with
the Lagrange multiplier $l$:
\begin{align}
\label{eq_J_pca}
J
&=
\frac{1}{2}\, \mb{w}^T \mb{C} \mb{w} - \frac{1}{2}\, l \mleft( \mb{w}^T
\mb{w} - 1 \mright).
\end{align}
Note that the sign of the constraint term is arbitrary. A negative sign
is used here such that the Lagrange parameter in the fixed points
coincides with an eigenvalue (see below) rather than a negative
eigenvalue.

Learning rules are derived from a Newton descent in $J$ over the
extended variable vector $\begin{pmatrix}\mb{w}^T & l\end{pmatrix}^T$
for a selected fixed point.

\section{Derivation of the General Hessian}\label{sec_hessian}

%
For the Newton descent, we have to determine the Hessian $\mb{H}$ and
the gradient of $J$. The first-order derivatives of $J$ are
\begin{align}
\label{eq_eigen}
\mleft( \frac{\partial}{\partial \mb{w}}J \mright)^T
&=
\mb{C} \mb{w} - l \mb{w}\\
\label{eq_constraint}
\mleft( \frac{\partial}{\partial l}J \mright)^T
&=
- \frac{1}{2}\, \mleft( \mb{w}^T \mb{w} - 1 \mright).
\end{align}
We see that equation \eqref{eq_eigen} is the eigen equation; together
with \eqref{eq_constraint} we obtain the fixed points $\mb{w} =
\mb{v}_{i}$ and $l = \lambda_{i}$, with $\mb{w}^T \mb{w} = 1$. We see
that the Lagrange multiplier coincides with the eigenvalues in the fixed
points.

The second-order derivatives are
\begin{align}
\frac{\partial}{\partial \mb{w}}\mleft( \frac{\partial}{\partial
\mb{w}}J \mright)^T
&=
\mb{C} - l \mb{I}_n\\
\frac{\partial}{\partial l}\mleft( \frac{\partial}{\partial \mb{w}}J
\mright)^T
&=
- \mb{w}\\
\frac{\partial}{\partial \mb{w}}\mleft( \frac{\partial}{\partial l}J
\mright)^T
&=
- \mb{w}^T\\
\frac{\partial}{\partial l}\mleft( \frac{\partial}{\partial l}J
\mright)^T
&=
0,
\end{align}
which gives the Hessian
\begin{align}
\label{eq_H}
\mb{H}
&=
\begin{pmatrix}\mb{C} - l \mb{I}_n & - \mb{w}\\- \mb{w}^T &
0\end{pmatrix}.
\end{align}
In both our original and the novel framework, learning rules are derived
by evaluating the Hessian at a chosen fixed point. This step is
performed by applying an orthogonal similarity transformation to the
Hessian, deriving approximations, inverting the matrix (as required for
the Newton descent), and transforming back. We choose the orthogonal
transformation matrix
\begin{align}
\label{eq_T}
\mb{T}
&=
\begin{pmatrix}\mb{V} & \mb{0}_n\\\mb{0}_n^T & 1\end{pmatrix}
\end{align}
and obtain the transformed Hessian
\begin{align}
&\phantom{{}={}} {\mb{H}^*} \nonumber \\
& = \mb{T}^T \mb{H} \mb{T}\\
& = \begin{pmatrix}\mb{V} & \mb{0}_n\\\mb{0}_n^T & 1\end{pmatrix}^T
\begin{pmatrix}\mb{C} - l \mb{I}_n & - \mb{w}\\- \mb{w}^T &
0\end{pmatrix} \begin{pmatrix}\mb{V} & \mb{0}_n\\\mb{0}_n^T &
1\end{pmatrix}\\
& = \begin{pmatrix}\mb{V}^T \mb{C} \mb{V} - \mb{V}^T l \mb{V} & - \mb{V}^T
\mb{w} \\\mleft( - \mb{w}^T \mright) \mb{V} & 0\end{pmatrix}\\
&\label{eq_Hstar_gen}
 = \begin{pmatrix}\bs{\Lambda} - l \mb{I}_n & \mb{V}^T \mleft( - \mb{w}
\mright) \\\mleft( - \mb{w}^T \mright) \mb{V} & 0\end{pmatrix}.
\end{align}
In the following, we choose the Hessian for the {\em principal}
eigenpair (section \ref{sec_principal}) and for an {\em arbitrary}
eigenpair (section \ref{sec_arbitrary}). This involves assumptions on
the relationships between eigenvalues. The Hessians are finally inverted
as required for the Newton descent. In the derivation, we approximate
the true but unknown eigenpair with the estimated eigenpair. Note that
this make it difficult to predict the behavior of the Newton descent far
from the desired fixed point.

\section{Stability Analysis for Exact Hessians}\label{sec_stability_exact}

%
In the following we analyze the effect of the approximations used in the
derivation of the inverse Hessian. We study the stability of fixed
points in a Newton descent for an {\em exact} inverse Hessian, for both
the desired fixed point and the undesired fixed points. Note that the
exact inverse Hessian wouldn't be available in an application. We
perform this analysis as a baseline since differences in behavior to
this case can then be explained as effect of the necessary
approximations.

In the following derivation we assume unordered eigenvectors
$\tilde{\mb{V}}$ and eigenvalues $\tilde{\bs{\Lambda}}$, both in
arbitrary order of eigenvalues. We assume a Newton descent is performed
for the Hessian matrix $\bar{\mb{H}}_{i}$ obtained at fixed point $i$
where $\mb{w} = \tilde{\mb{v}}_{i}$ and $l = \tilde{\lambda}_{i}$:
\begin{align}
\begin{pmatrix}\dot{\mb{w}}\\\dot{l}\end{pmatrix}
&=
- \bar{\mb{H}}_{i}^{-1} \begin{pmatrix}\mleft( \frac{\partial}{\partial
\mb{w}}J \mright)^T\\\mleft( \frac{\partial}{\partial l}J
\mright)^T\end{pmatrix}.
\end{align}
The Jacobian of this system is evaluated at the fixed point $j$ where
$\mb{w} = \tilde{\mb{v}}_{j}$ and $l = \tilde{\lambda}_{j}$:
\begin{align}
&\phantom{{}={}} \bar{\mb{J}} \nonumber \\
& = \begin{pmatrix}\frac{\partial}{\partial \mb{w}}\dot{\mb{w}} &
\frac{\partial}{\partial l}\dot{\mb{w}}\\\frac{\partial}{\partial
\mb{w}}\dot{l} & \frac{\partial}{\partial l}\dot{l}\end{pmatrix}\\
& = - \bar{\mb{H}}_{i}^{-1} \bar{\mb{H}}_{j}.
\end{align}
For $i = j$, i.e. for the case where the inverse Hessian corresponds to
the fixed point under consideration, we get $\mb{J} = - \mb{I}_{n+1}$.
We see that the system converges with the same speed from all directions
and the speed is constant ($-1$).

For $i \neq j$, i.e. for the case where the inverse Hessian belongs to a
different fixed point, the eigenvalues of the Jacobian determine
stability at undesired fixed point.

Since it is probably challenging to obtain a general statement on the
stability for an arbitrary criterion and an arbitrary constraint, we
only study the case of PCA.

We first apply an orthogonality transformation with matrix
\begin{align}
\tilde{\mb{T}}
&=
\begin{pmatrix}\tilde{\mb{V}} & \mb{0}_n\\\mb{0}_n^T & 1\end{pmatrix}
\end{align}
to the Jacobian:
\begin{align}
&\phantom{{}={}} {\bar{\mb{J}}^*} \nonumber \\
& = \tilde{\mb{T}}^T \bar{\mb{J}} \tilde{\mb{T}}\\
& = \tilde{\mb{T}}^T \mleft( - \bar{\mb{H}}_{i}^{-1} \bar{\mb{H}}_{j}
\mright) \tilde{\mb{T}}\\
& = - \tilde{\mb{T}}^T \bar{\mb{H}}_{i}^{-1} \tilde{\mb{T}}
\tilde{\mb{T}}^{-1} \bar{\mb{H}}_{j} \tilde{\mb{T}}\\
& = - \mleft( \tilde{\mb{T}}^T \bar{\mb{H}}_{i} \tilde{\mb{T}} \mright)^{-1}
\tilde{\mb{T}}^{-1} \bar{\mb{H}}_{j} \tilde{\mb{T}}\\
& = - {{} {{\bar{\mb{H}}^*}_{i}}}^{-1} {{} {{\bar{\mb{H}}^*}_{j}}}.
\end{align}
The orthogonal similarity transformation of $\bar{\mb{J}}$ doesn't its
affect eigenvalue spectrum, so we analyze the eigenvalues of transformed
Jacobian ${\bar{\mb{J}}^*}$ instead:
\begin{align}
\label{eq_Jtrans}
{\bar{\mb{J}}^*}
&=
- {{} {{\bar{\mb{H}}^*}_{i}}}^{-1} {{} {{\bar{\mb{H}}^*}_{j}}}.
\end{align}
We start by determining the Hessian for fixed point $i$ where $\mb{w} =
\tilde{\mb{v}}_{i}$ and $l = \tilde{\lambda}_{i}$. We take the general
Hessian from \eqref{eq_H}
\begin{align}
\mb{H}
&=
\begin{pmatrix}\mb{C} - l \mb{I}_n & - \mb{w}\\- \mb{w}^T &
0\end{pmatrix},
\end{align}
determine the Hessian after orthogonal similarity transformation
analogous to \eqref{eq_Hstar_gen}
\begin{align}
&\phantom{{}={}} {\mb{H}^*} \nonumber \\
& = \tilde{\mb{T}}^T \mb{H} \tilde{\mb{T}}\\
& = \begin{pmatrix}\tilde{\mb{V}} & \mb{0}_n\\\mb{0}_n^T & 1\end{pmatrix}^T
\mb{H} \begin{pmatrix}\tilde{\mb{V}} & \mb{0}_n\\\mb{0}_n^T &
1\end{pmatrix}\\
& = \begin{pmatrix}\tilde{\mb{V}} & \mb{0}_n\\\mb{0}_n^T & 1\end{pmatrix}^T
\begin{pmatrix}\mb{C} - l \mb{I}_n & - \mb{w}\\- \mb{w}^T &
0\end{pmatrix} \begin{pmatrix}\tilde{\mb{V}} & \mb{0}_n\\\mb{0}_n^T &
1\end{pmatrix}\\
& = \begin{pmatrix}\tilde{\bs{\Lambda}} - l \mb{I}_n & \tilde{\mb{V}}^T
\mleft( - \mb{w} \mright) \\\mleft( - \mb{w}^T \mright) \tilde{\mb{V}} &
0\end{pmatrix},
\end{align}
and insert fixed point $i$
\begin{align}
&\phantom{{}={}} {{} {{\bar{\mb{H}}^*}_{i}}} \nonumber \\
& = \begin{pmatrix}\tilde{\bs{\Lambda}} - \tilde{\lambda}_{i} \mb{I}_n &
\tilde{\mb{V}}^T \mleft( - \tilde{\mb{v}}_{i} \mright) \\\mleft( -
\tilde{\mb{v}}_{i}^T \mright) \tilde{\mb{V}} & 0\end{pmatrix}\\
& = \begin{pmatrix}\tilde{\bs{\Lambda}} - \tilde{\lambda}_{i} \mb{I}_n & -
\mb{e}_{i}\\\mleft( - \mb{e}_{i} \mright)^T & 0\end{pmatrix}\\
& = \begin{pmatrix}\tilde{\lambda}_{1} - \tilde{\lambda}_{i} & & & & & 0\\ &
\ddots & & & & \vdots \\ & & \tilde{\lambda}_{i} - \tilde{\lambda}_{i} &
& & - 1\\ & & & \ddots & & \vdots \\ & & & & \tilde{\lambda}_{n} -
\tilde{\lambda}_{i} & 0\\0 & \cdots & - 1 & \cdots & 0 & 0\end{pmatrix}\\
& = \begin{pmatrix}\tilde{\lambda}_{1} - \tilde{\lambda}_{i} & & & & & 0\\ &
\ddots & & & & \vdots \\ & & 0 & & & - 1\\ & & & \ddots & & \vdots \\ &
& & & \tilde{\lambda}_{n} - \tilde{\lambda}_{i} & 0\\0 & \cdots & - 1 &
\cdots & 0 & 0\end{pmatrix}\\
& = \begin{pmatrix}\tilde{\lambda}_{1} - \tilde{\lambda}_{i} & & & & & 0\\ &
\ddots & & & & \vdots \\ & & - 1 & & & 0\\ & & & \ddots & & \vdots \\ &
& & & \tilde{\lambda}_{n} - \tilde{\lambda}_{i} & 0\\0 & \cdots & 0 &
\cdots & 0 & - 1\end{pmatrix} \mb{P}_{i,n+1}.
\end{align}
We then invert the transformed Hessian, and exploit the property
$\mb{P}_{i,n+1}^{-1} = \mb{P}_{i,n+1}$ of the elementary permutation:
\begin{align}
&\phantom{{}={}} {{} {{\bar{\mb{H}}^*}_{i}}}^{-1} \nonumber \\
& = \mb{P}_{i,n+1} \begin{pmatrix}\mleft( \tilde{\lambda}_{1} -
\tilde{\lambda}_{i} \mright)^{-1} & & & & & 0\\ & \ddots & & & & \vdots
\\ & & - 1 & & & 0\\ & & & \ddots & & \vdots \\ & & & & \mleft(
\tilde{\lambda}_{n} - \tilde{\lambda}_{i} \mright)^{-1} & 0\\0 & \cdots
& 0 & \cdots & 0 & - 1\end{pmatrix}\\
& = \begin{pmatrix}\mleft( \tilde{\lambda}_{1} - \tilde{\lambda}_{i}
\mright)^{-1} & & & & & 0\\ & \ddots & & & & \vdots \\ & & 0 & & & - 1\\
& & & \ddots & & \vdots \\ & & & & \mleft( \tilde{\lambda}_{n} -
\tilde{\lambda}_{i} \mright)^{-1} & 0\\0 & \cdots & - 1 & \cdots & 0 &
0\end{pmatrix}.
\end{align}
In the next step, we insert the inverse transformed Hessian into
\eqref{eq_Jtrans}, replace the diagonal matrices in the upper-left
corner by $\mb{A}_{i}$ and $\mb{B}_{j}$, respectively, and assume $i
\neq j$:
\begin{align}
&\phantom{{}={}} {\bar{\mb{J}}^*} \nonumber \\
& = - {{} {{\bar{\mb{H}}^*}_{i}}}^{-1} {{} {{\bar{\mb{H}}^*}_{j}}}\\
& = \resizebox{\minof{\width}{\symaeqwidth}}{!}{%
$- \begin{pmatrix}\mleft( \tilde{\lambda}_{1} - \tilde{\lambda}_{i}
\mright)^{-1} & & & & & 0\\ & \ddots & & & & \vdots \\ & & 0 & & & - 1\\
& & & \ddots & & \vdots \\ & & & & \mleft( \tilde{\lambda}_{n} -
\tilde{\lambda}_{i} \mright)^{-1} & 0\\0 & \cdots & - 1 & \cdots & 0 &
0\end{pmatrix} \begin{pmatrix}\tilde{\lambda}_{1} - \tilde{\lambda}_{j}
& & & & & 0\\ & \ddots & & & & \vdots \\ & & 0 & & & - 1\\ & & & \ddots
& & \vdots \\ & & & & \tilde{\lambda}_{n} - \tilde{\lambda}_{j} & 0\\0 &
\cdots & - 1 & \cdots & 0 & 0\end{pmatrix} $%
}\\
& = - \begin{pmatrix}\mb{A}_{i} & - \mb{e}_{i}\\- \mb{e}_{i}^T &
0\end{pmatrix} \begin{pmatrix}\mb{B}_{j} & - \mb{e}_{j}\\- \mb{e}_{j}^T
& 0\end{pmatrix}\\
& = - \begin{pmatrix}\mb{A}_{i} \mb{B}_{j} + \mb{e}_{i} \mb{e}_{j}^T & -
\mb{A}_{i} \mb{e}_{j} \\- \mb{e}_{i}^T \mb{B}_{j} & \mb{e}_{i}^T
\mb{e}_{j} \end{pmatrix}\\
& = - \begin{pmatrix}\mb{A}_{i} \mb{B}_{j} + \mb{e}_{i} \mb{e}_{j}^T & -
\mleft( \tilde{\lambda}_{j} - \tilde{\lambda}_{i} \mright)^{-1}
\mb{e}_{j} \\- \mb{e}_{i}^T \mleft( \tilde{\lambda}_{i} -
\tilde{\lambda}_{j} \mright) & 0\end{pmatrix}.
\end{align}
Since any ordering of eigenpairs is possible, we can chose arbitrary
indices, $i = 1$, $j = 2$:
\begin{align}
&\phantom{{}={}} {\bar{\mb{J}}^*} \nonumber \\
& = \resizebox{\minof{\width}{\symaeqwidth}}{!}{%
$- \begin{pmatrix}0 & & & & & - 1\\ & \mleft( \tilde{\lambda}_{2} -
\tilde{\lambda}_{1} \mright)^{-1} & & & & 0\\ & & \mleft(
\tilde{\lambda}_{3} - \tilde{\lambda}_{1} \mright)^{-1} & & & 0\\ & & &
\ddots & & \vdots \\ & & & & \mleft( \tilde{\lambda}_{n} -
\tilde{\lambda}_{1} \mright)^{-1} & 0\\- 1 & 0 & 0 & \cdots & 0 &
0\end{pmatrix} \begin{pmatrix}\tilde{\lambda}_{1} - \tilde{\lambda}_{2}
& & & & & 0\\ & 0 & & & & - 1\\ & & \tilde{\lambda}_{3} -
\tilde{\lambda}_{2} & & & 0\\ & & & \ddots & & \vdots \\ & & & &
\tilde{\lambda}_{n} - \tilde{\lambda}_{2} & 0\\0 & - 1 & 0 & \cdots & 0
& 0\end{pmatrix} $%
}\\
& = \resizebox{\minof{\width}{\symaeqwidth}}{!}{%
$- \begin{pmatrix}0 & 1 & & & & 0\\0 & 0 & & & & - \mleft(
\tilde{\lambda}_{2} - \tilde{\lambda}_{1} \mright)^{-1}\\ & & \mleft(
\tilde{\lambda}_{3} - \tilde{\lambda}_{1} \mright)^{-1} \mleft(
\tilde{\lambda}_{3} - \tilde{\lambda}_{2} \mright) & & & 0\\ & & &
\ddots & & \vdots \\ & & & & \mleft( \tilde{\lambda}_{n} -
\tilde{\lambda}_{1} \mright)^{-1} \mleft( \tilde{\lambda}_{n} -
\tilde{\lambda}_{2} \mright) & 0\\- \mleft( \tilde{\lambda}_{1} -
\tilde{\lambda}_{2} \mright) & 0 & 0 & \cdots & 0 & 0\end{pmatrix}$%
}\\
& = \resizebox{\minof{\width}{\symaeqwidth}}{!}{%
$\begin{pmatrix}0 & - 1 & & & & 0\\0 & 0 & & & & \mleft(
\tilde{\lambda}_{2} - \tilde{\lambda}_{1} \mright)^{-1}\\ & & - \mleft(
\tilde{\lambda}_{3} - \tilde{\lambda}_{1} \mright)^{-1} \mleft(
\tilde{\lambda}_{3} - \tilde{\lambda}_{2} \mright) & & & 0\\ & & &
\ddots & & \vdots \\ & & & & - \mleft( \tilde{\lambda}_{n} -
\tilde{\lambda}_{1} \mright)^{-1} \mleft( \tilde{\lambda}_{n} -
\tilde{\lambda}_{2} \mright) & 0\\\tilde{\lambda}_{1} -
\tilde{\lambda}_{2} & 0 & 0 & \cdots & 0 & 0\end{pmatrix}$%
}.
\end{align}
We abbreviate $d_{i,j} = \tilde{\lambda}_{i} - \tilde{\lambda}_{j}$ and
get
\begin{align}
&\phantom{{}={}} {\bar{\mb{J}}^*} \nonumber \\
& = \begin{pmatrix}0 & - 1 & & & & 0\\0 & 0 & & & & d_{2,1}^{-1}\\ & & -
d_{3,1}^{-1} d_{3,2} & & & 0\\ & & & \ddots & & \vdots \\ & & & & -
d_{n,1}^{-1} d_{n,2} & 0\\- d_{2,1} & 0 & 0 & \cdots & 0 &
0\end{pmatrix}.
\end{align}
We determine the eigenvalues $\alpha$ of ${\bar{\mb{J}}^*}$ from
\begin{align}
&\phantom{{}={}} \opnl{det} \mleft\{ {\bar{\mb{J}}^*} - \alpha \mb{I}_{n+1} \mright\} \nonumber \\
& = \begin{vmatrix}- \alpha & - 1 & & & & 0\\0 & - \alpha & & & &
d_{2,1}^{-1}\\ & & \mleft( - d_{3,1}^{-1} d_{3,2} \mright) - \alpha & &
& 0\\ & & & \ddots & & \vdots \\ & & & & \mleft( - d_{n,1}^{-1} d_{n,2}
\mright) - \alpha & 0\\- d_{2,1} & 0 & 0 & \cdots & 0 & -
\alpha\end{vmatrix}_{n + 1}\\
\shortintertext{develop along the upper row,}
& = \mleft( - \alpha \mright) \begin{vmatrix}- \alpha & & & & d_{2,1}^{-1}\\
& \mleft( - d_{3,1}^{-1} d_{3,2} \mright) - \alpha & & & 0\\ & & \ddots
& & \vdots \\ & & & \mleft( - d_{n,1}^{-1} d_{n,2} \mright) - \alpha &
0\\0 & 0 & \cdots & 0 & - \alpha\end{vmatrix}_{n}
\\&\phantom{{}={}}
\nonumber
{} \! + \begin{vmatrix}0 & & & & d_{2,1}^{-1}\\ & \mleft( - d_{3,1}^{-1}
d_{3,2} \mright) - \alpha & & & 0\\ & & \ddots & & \vdots \\ & & &
\mleft( - d_{n,1}^{-1} d_{n,2} \mright) - \alpha & 0\\- d_{2,1} & 0 &
\cdots & 0 & - \alpha\end{vmatrix}_{n}\\
\shortintertext{develop the first determinant along the left column and the second determinant along the first row,}
& = \alpha^2 \begin{vmatrix}\mleft( - d_{3,1}^{-1} d_{3,2} \mright) - \alpha
& & & 0\\ & \ddots & & \vdots \\ & & \mleft( - d_{n,1}^{-1} d_{n,2}
\mright) - \alpha & 0\\0 & \cdots & 0 & - \alpha\end{vmatrix}_{n - 1}
\\&\phantom{{}={}}
\nonumber
{} \! + d_{2,1}^{-1} \mleft( - 1 \mright)^{1 + n} \begin{vmatrix}0 &
\mleft( - d_{3,1}^{-1} d_{3,2} \mright) - \alpha & \cdots & 0\\\vdots &
& \ddots & \vdots \\0 & & & \mleft( - d_{n,1}^{-1} d_{n,2} \mright) -
\alpha\\- d_{2,1} & 0 & \cdots & 0\end{vmatrix}_{n - 1}\\
\shortintertext{develop the first determinant along the last column, and the second determinant along the first column,}
& = \mleft( - \alpha^{3} \mright) \mleft( - 1 \mright)^{2 n - 2}
\begin{vmatrix}\mleft( - d_{3,1}^{-1} d_{3,2} \mright) - \alpha & & \\ &
\ddots & \\ & & \mleft( - d_{n,1}^{-1} d_{n,2} \mright) -
\alpha\end{vmatrix}_{n - 2}
\\&\phantom{{}={}}
\nonumber
{} \! + d_{2,1}^{-1} \mleft( - d_{2,1} \mright) \mleft( - 1 \mright)^{1
+ n} \mleft( - 1 \mright)^{n} \begin{vmatrix}\mleft( - d_{3,1}^{-1}
d_{3,2} \mright) - \alpha & & \\ & \ddots & \\ & & \mleft( -
d_{n,1}^{-1} d_{n,2} \mright) - \alpha\end{vmatrix}_{n - 2}\\
\shortintertext{and finally factor out}
& = \mleft( \mleft[ - \alpha^{3} \mright] + 1 \mright)
\begin{vmatrix}\mleft( - d_{3,1}^{-1} d_{3,2} \mright) - \alpha & & \\ &
\ddots & \\ & & \mleft( - d_{n,1}^{-1} d_{n,2} \mright) -
\alpha\end{vmatrix}_{n - 2}\\
& = 0.
\end{align}
The eigenvalues are $1$, $-\frac{1}{2} \pm \frac{\sqrt{3}}{2} i$ as
complex solutions of $\alpha^3 = 1$, and
\begin{align}
\label{eq_stab_exact_alphak}
\alpha_{k} & = - d_{k,1}^{-1} d_{k,2} = - \mleft( \tilde{\lambda}_{k} - \tilde{\lambda}_{1} \mright)^{-1}
\mleft( \tilde{\lambda}_{k} - \tilde{\lambda}_{2} \mright)
\quad\mbox{for}\;k=3,\ldots,n.
\end{align}
The eigenvalues were confirmed in numerical experiments. The eigenvalue
of $1$ and the two complex eigenvalues with negative real part of
$-\frac{1}{2}$ indicate that we have a saddle point if the inverse
Hessian doesn't correspond to the fixed point. Surprisingly, this
statement is independent of the eigenvalue spectrum.\footnote{If we look
at \eqref{eq_stab_exact_alphak} in addition, we see that positive
(instable) eigenvalues are introduced for any $\tilde{\lambda}_{k}$
lying between $\tilde{\lambda}_{1}$ and $\tilde{\lambda}_{2}$. All
eigenvalues in \eqref{eq_stab_exact_alphak} are negative if no such
$\tilde{\lambda}_{k}$ can be found. The implications are currently
unclear.}

We conclude that at least if the Hessians would be known exactly (which
is not the case when the learning rules are applied), there is only a
single stable fixed point at the desired location, whereas all other
fixed points are saddle points, thus the system will converge to the
desired fixed point.

However, since the derivations of the learning rules below necessarily
use {\em approximated} Hessians, the stability of each learning rule
needs to be studied. Moreover, the learning rules may sometimes converge
to an additional set of fixed points which is introduced by Newton's
method.\footnote{It is presently not clear why we obtain eigenvalues of
the Jacobian with imaginary part for exact Hessians, but purely real
eigenvalues for the approximated Hessians.}

\section{Principal Eigenpair}\label{sec_principal}

%
In the following we derive coupled learning rules for the {\em
principal} eigenpair (the one with largest eigenvalue). The resulting
learning rules are the same as in our previous publication
\citep[][]{own_Moeller04a}, but are now obtained from the
Lagrange-Newton approach.

\subsection{Inverse Hessian}

%
$\bs{\Lambda}$ is the matrix of all eigenvalues of $\mb{C}$, sorted in
descending order along the main diagonal. We assume that the eigenvalues
are distinct (i.e. pairwise different). The orthogonal matrix $\mb{V}$
is the matrix of all eigenvectors of $\mb{C}$; the eigenvectors in its
columns appear in the same order as the eigenvalues. We want to
determine the principal eigenpair $i = 1$, i.e. the one with the largest
eigenvalue. In the vicinity of the desired fixed point for $i = 1$ we
have $l \approx \lambda_{1}$ and $\mb{w} \approx \mb{v}_{1}$, such that
\eqref{eq_Hstar_gen} turns into:
\begin{align}
&\phantom{{}={}} {\mb{H}^*} \nonumber \\
& \approx \begin{pmatrix}\bs{\Lambda} - \lambda_{1} \mb{I}_n & \mb{V}^T \mleft( -
\mb{v}_{1} \mright) \\\mleft( - \mb{v}_{1}^T \mright) \mb{V} &
0\end{pmatrix}\\
& = \begin{pmatrix}\bs{\Lambda} - \lambda_{1} \mb{I}_n & - \mb{e}_{1}\\-
\mb{e}_{1}^T & 0\end{pmatrix}.
\end{align}
We see that element $i$ of the diagonal matrix $\bs{\Lambda} -
\lambda_{1} \mb{I}_n $ is $\lambda_{i} - \lambda_{1}$. We assume
$\lambda_{1} \gg \lambda_{i}$ for $i > 1$ and therefore $\lambda_{i} -
\lambda_{1} \approx - \lambda_{1}$; the first diagonal element is
$0$:\footnote{This approximation attempts to eliminate the dependency on
later eigenpair estimates ($i > 1$) which are unknown from the
perspective of the principal-component estimator. However, it may be
interesting to explore a derivation where the approximation is not
applied.}
\begin{align}
&\phantom{{}={}} {\mb{H}^*} \nonumber \\
& \approx \begin{pmatrix}\lambda_{1} \mleft( \mb{e}_{1} \mb{e}_{1}^T - \mb{I}_n
\mright) & - \mb{e}_{1}\\- \mb{e}_{1}^T & 0\end{pmatrix}\\
& \approx \begin{pmatrix}l \mleft( \mb{e}_{1} \mb{e}_{1}^T - \mb{I}_n \mright) & -
\mb{e}_{1}\\- \mb{e}_{1}^T & 0\end{pmatrix}\\
& = - \begin{pmatrix}0 & \mb{0}_{n-1}^T & 1\\\mb{0}_{n-1} & l \mb{I}_{n-1} &
\mb{0}_{n-1}\\1 & \mb{0}_{n-1}^T & 0\end{pmatrix}\\
& = - \begin{pmatrix}1 & \mb{0}_{n-1}^T & 0\\\mb{0}_{n-1} & l \mb{I}_{n-1} &
\mb{0}_{n-1}\\0 & \mb{0}_{n-1}^T & 1\end{pmatrix} \mb{P}_{1,n+1}.
\end{align}
In the last step we exchanged column $1$ and $n+1$ by post-multiplying
with the corresponding elementary permutation matrix $\mb{P}_{1,n+1}$.
Note that we have $\mb{P}_{1,n+1} = \mb{P}_{1,n+1}^{-1}$ since we have
an elementary permutation matrix which only exchanges two columns. The
resulting matrix is diagonal and therefore easy to invert; this is a
more concise derivation than in our previous work
\citep[][]{own_Moeller04a,own_Moeller20}. After the inversion we obtain
a pre-multiplication by $\mb{P}_{1,n+1}$ which performs a permutation of
rows:
\begin{align}
&\phantom{{}={}} {\mb{H}^*}^{-1} \nonumber \\
& \approx - \mb{P}_{1,n+1} \begin{pmatrix}1 & \mb{0}_{n-1}^T & 0\\\mb{0}_{n-1} &
l^{-1} \mb{I}_{n-1} & \mb{0}_{n-1}\\0 & \mb{0}_{n-1}^T & 1\end{pmatrix}\\
& = - \begin{pmatrix}0 & \mb{0}_{n-1}^T & 1\\\mb{0}_{n-1} & l^{-1}
\mb{I}_{n-1} & \mb{0}_{n-1}\\1 & \mb{0}_{n-1}^T & 0\end{pmatrix}\\
& = \begin{pmatrix}l^{-1} \mleft( \mb{e}_{1} \mb{e}_{1}^T - \mb{I}_n
\mright) & - \mb{e}_{1}\\- \mb{e}_{1}^T & 0\end{pmatrix}.
\end{align}
For the transformation back to $\mb{H}^{-1}$ we apply
\begin{align}
{\mb{H}^*}
&=
\mb{T}^T \mb{H} \mb{T}\\
{\mb{H}^*}^{-1}
&=
\mleft( \mb{T}^T \mb{H} \mb{T} \mright)^{-1}\\
{\mb{H}^*}^{-1}
&=
\mb{T}^T \mb{H}^{-1} \mb{T}\\
\mb{H}^{-1}
&=
\mb{T} {\mb{H}^*}^{-1} \mb{T}^T.
\end{align}
We multiply the block matrices
\begin{align}
&\phantom{{}={}} \mb{H}^{-1} \nonumber \\
& \approx \mb{T} {\mb{H}^*}^{-1} \mb{T}^T\\
& = \begin{pmatrix}\mb{V} & \mb{0}_n\\\mb{0}_n^T & 1\end{pmatrix}
\begin{pmatrix}l^{-1} \mleft( \mb{e}_{1} \mb{e}_{1}^T - \mb{I}_n
\mright) & - \mb{e}_{1}\\- \mb{e}_{1}^T & 0\end{pmatrix}
\begin{pmatrix}\mb{V} & \mb{0}_n\\\mb{0}_n^T & 1\end{pmatrix}^T\\
& = \begin{pmatrix}\mb{V} l^{-1} \mleft( \mb{e}_{1} \mb{e}_{1}^T -
\mb{I}_{n} \mright) \mb{V}^T & - \mb{V} \mb{e}_{1} \\\mleft( -
\mb{e}_{1}^T \mright) \mb{V}^T & 0\end{pmatrix}.
\end{align}
In the vicinity of the fixed point we have $\mb{V} \mb{e}_{1} =
\mb{v}_{1}$ and $\mb{v}_{1} \approx \mb{w}$, so we get
\begin{align}
&\phantom{{}={}} \mb{H}^{-1} \nonumber \\
&\label{eq_Hinv_principal}
 \approx \begin{pmatrix}l^{-1} \mleft( \mb{w} \mb{w}^T - \mb{I}_n \mright) & -
\mb{w}\\- \mb{w}^T & 0\end{pmatrix}.
\end{align}

\subsection{Newton Descent}\label{sec_newton_principal}

%
Now we insert the inverted Hessian \eqref{eq_Hinv_principal} into the
Newton descent:
\begin{align}
&\phantom{{}={}} \begin{pmatrix}\dot{\mb{w}}\\\dot{l}\end{pmatrix} \nonumber \\
& = - \mb{H}^{-1} \begin{pmatrix}\mleft( \frac{\partial}{\partial \mb{w}}J
\mright)^T\\\mleft( \frac{\partial}{\partial l}J \mright)^T\end{pmatrix}\\
& \approx - \begin{pmatrix}l^{-1} \mleft( \mb{w} \mb{w}^T - \mb{I}_n \mright) & -
\mb{w}\\- \mb{w}^T & 0\end{pmatrix} \begin{pmatrix}\mb{C} \mb{w} - l
\mb{w} \\- \frac{1}{2}\, \mleft( \mb{w}^T \mb{w} - 1 \mright)
\end{pmatrix}\\
& = - \begin{pmatrix}l^{-1} \mleft( \mb{w} \mb{w}^T - \mb{I}_n \mright)
\mb{C} \mb{w} - l^{-1} \mleft( \mb{w} \mb{w}^T - \mb{I}_n \mright) l
\mb{w} + \mb{w} \frac{1}{2}\, \mleft( \mb{w}^T \mb{w} - 1 \mright)
\\\mleft( - \mb{w}^T \mb{C} \mb{w} \mright) + \mb{w}^T l \mb{w}
\end{pmatrix}\\
& = \begin{pmatrix}\mleft( - l^{-1} \mb{w} \mb{w}^T \mb{C} \mb{w} \mright) +
l^{-1} \mb{C} \mb{w} + l^{-1} l \mb{w} \mb{w}^T \mb{w} - l^{-1} l \mb{w}
- \frac{1}{2}\, \mb{w} \mb{w}^T \mb{w} + \frac{1}{2}\, \mb{w} \\\mb{w}^T
\mb{C} \mb{w} - l \mb{w}^T \mb{w} \end{pmatrix}\\
& = \begin{pmatrix}\mleft( - l^{-1} \mb{w} \mb{w}^T \mb{C} \mb{w} \mright) +
l^{-1} \mb{C} \mb{w} + \mb{w} \mb{w}^T \mb{w} - \mb{w} - \frac{1}{2}\,
\mb{w} \mb{w}^T \mb{w} + \frac{1}{2}\, \mb{w} \\\mb{w}^T \mb{C} \mb{w} -
l \mb{w}^T \mb{w} \end{pmatrix}\\
& = \begin{pmatrix}\mleft( - l^{-1} \mb{w} \mb{w}^T \mb{C} \mb{w} \mright) +
l^{-1} \mb{C} \mb{w} + \frac{1}{2}\, \mb{w} \mb{w}^T \mb{w} -
\frac{1}{2}\, \mb{w} \\\mb{w}^T \mb{C} \mb{w} - l \mb{w}^T \mb{w}
\end{pmatrix}\\
&\label{eq_newton_principal}
 = \begin{pmatrix}l^{-1} \mleft( \mb{C} \mb{w} - \mleft[ \mb{w}^T \mb{C}
\mb{w} \mright] \mb{w} \mright) + \frac{1}{2}\, \mleft( \mb{w}^T \mb{w}
- 1 \mright) \mb{w} \\\mb{w}^T \mb{C} \mb{w} - l \mb{w}^T \mb{w}
\end{pmatrix}.
\end{align}
These equations are the same as the ``nPCA'' system derived in our
previous work \citep[][]{own_Moeller04a}.

There is one observation to report: Applying the Newton descent
introduces additional fixed points which are not present in the original
set of fixed points. The differential equation above obviously has a
fixed point at $\mb{w} = \mb{0}_n$ for arbitrary $l$ ($l \neq 0$). These
additional fixed points probably appear since the Hessian is singular
for these values.

\subsection{Stability Analysis}

%
We now analyse the stability of the coupled learning rule
\eqref{eq_newton_principal}:
\begin{align}
\begin{pmatrix}\dot{\mb{w}}\\\dot{l}\end{pmatrix}
&=
\begin{pmatrix}l^{-1} \mleft( \mb{C} \mb{w} - \mleft[ \mb{w}^T \mb{C}
\mb{w} \mright] \mb{w} \mright) + \frac{1}{2}\, \mleft( \mb{w}^T \mb{w}
- 1 \mright) \mb{w} \\\mb{w}^T \mb{C} \mb{w} - l \mb{w}^T \mb{w}
\end{pmatrix}.
\end{align}
Stability at a given fixed point is evident from the eigenvalues of the
Jacobian evaluated at this fixed point. We apply derivatives from
appendix \ref{app_derivatives} and get the Jacobian
\begin{align}
&\phantom{{}={}} \mb{J} \nonumber \\
& = \begin{pmatrix}\frac{\partial}{\partial \mb{w}}\dot{\mb{w}} &
\frac{\partial}{\partial l}\dot{\mb{w}}\\\frac{\partial}{\partial
\mb{w}}\dot{l} & \frac{\partial}{\partial l}\dot{l}\end{pmatrix}\\
& = \resizebox{\minof{\width}{\symaeqwidth}}{!}{%
$\begin{pmatrix}l^{-1} \mleft( \mb{C} - 2 \mb{w} \mb{w}^T \mb{C} -
\mleft[ \mb{w}^T \mb{C} \mb{w} \mright] \mb{I}_n \mright) + \mb{w}
\mb{w}^T + \frac{1}{2}\, \mleft( \mb{w}^T \mb{w} \mright) \mb{I}_n -
\frac{1}{2}\, \mb{I}_n & - l^{-2} \mleft( \mb{C} \mb{w} - \mleft[
\mb{w}^T \mb{C} \mb{w} \mright] \mb{w} \mright) \\2 \mleft( \mb{w}^T
\mb{C} - l \mb{w}^T \mright) & - \mb{w}^T \mb{w} \end{pmatrix}$%
}.
\end{align}
Now we insert fixed point $q$ specified by $\begin{pmatrix}\mb{v}_{q}^T
& \lambda_{q}\end{pmatrix}^T$ with $\mb{v}_{q}^T \mb{v}_{q} = 1$ and
$\mb{C} \mb{v}_{q} = \lambda_{q} \mb{v}_{q} $:
\begin{align}
&\phantom{{}={}} \bar{\mb{J}} \nonumber \\
& = \mleft.\mb{J}\mright|_{\mb{w} = \mb{v}_{q},\; l = \lambda_{q}}\\
& = \resizebox{\minof{\width}{\symaeqwidth}}{!}{%
$\begin{pmatrix}\lambda_{q}^{-1} \mleft( \mb{C} - 2 \mb{v}_{q}
\mb{v}_{q}^T \mb{C} - \mleft[ \mb{v}_{q}^T \mb{C} \mb{v}_{q} \mright]
\mb{I}_n \mright) + \mb{v}_{q} \mb{v}_{q}^T + \frac{1}{2}\, \mleft(
\mb{v}_{q}^T \mb{v}_{q} \mright) \mb{I}_n - \frac{1}{2}\, \mb{I}_n & -
\lambda_{q}^{-2} \mleft( \mb{C} \mb{v}_{q} - \mleft[ \mb{v}_{q}^T \mb{C}
\mb{v}_{q} \mright] \mb{v}_{q} \mright) \\2 \mleft( \mb{v}_{q}^T \mb{C}
- \lambda_{q} \mb{v}_{q}^T \mright) & - \mb{v}_{q}^T \mb{v}_{q}
\end{pmatrix}$%
}\\
& = \resizebox{\minof{\width}{\symaeqwidth}}{!}{%
$\begin{pmatrix}\lambda_{q}^{-1} \mleft( \mb{C} - 2 \mb{v}_{q}
\mb{v}_{q}^T \lambda_{q} - \lambda_{q} \mb{I}_n \mright) + \mb{v}_{q}
\mb{v}_{q}^T & \mb{0}_n\\\mb{0}_n^T & - 1\end{pmatrix}$%
}\\
& = \resizebox{\minof{\width}{\symaeqwidth}}{!}{%
$\begin{pmatrix}\lambda_{q}^{-1} \mb{C} - 2 \mb{v}_{q} \mb{v}_{q}^T -
\mb{I}_n + \mb{v}_{q} \mb{v}_{q}^T & \mb{0}_n\\\mb{0}_n^T & -
1\end{pmatrix}$%
}\\
& = \resizebox{\minof{\width}{\symaeqwidth}}{!}{%
$\begin{pmatrix}\lambda_{q}^{-1} \mb{C} - \mb{v}_{q} \mb{v}_{q}^T -
\mb{I}_n & \mb{0}_n\\\mb{0}_n^T & - 1\end{pmatrix}$%
}.
\end{align}
To extract the eigenvalues of $\mleft.\bar{\mb{J}}\mright|_{q}$, we
apply an orthogonal similarity transformation with $\mb{T}$ from
\eqref{eq_T}. Eigenvalues are invariant under a similarity
transformation. We obtain the transformed Jacobian ${\bar{\mb{J}}^*}$ at
the fixed point $q$:
\begin{align}
&\phantom{{}={}} {\bar{\mb{J}}^*} \nonumber \\
& = \begin{pmatrix}\mb{V} & \mb{0}_n\\\mb{0}_n^T & 1\end{pmatrix}^T
\bar{\mb{J}} \begin{pmatrix}\mb{V} & \mb{0}_n\\\mb{0}_n^T &
1\end{pmatrix}\\
& = \begin{pmatrix}\mb{V}^T & \mb{0}_n\\\mb{0}_n^T & 1\end{pmatrix}
\begin{pmatrix}\lambda_{q}^{-1} \mb{C} - \mb{v}_{q} \mb{v}_{q}^T -
\mb{I}_n & \mb{0}_n\\\mb{0}_n^T & - 1\end{pmatrix} \begin{pmatrix}\mb{V}
& \mb{0}_n\\\mb{0}_n^T & 1\end{pmatrix}\\
& = \begin{pmatrix}\mb{V}^T \lambda_{q}^{-1} \mb{C} \mb{V} - \mb{V}^T
\mb{v}_{q} \mb{v}_{q}^T \mb{V} - \mb{V}^T \mb{V} &
\mb{0}_{n}\\\mb{0}_{n}^T & - 1\end{pmatrix}\\
& = \begin{pmatrix}\lambda_{q}^{-1} \bs{\Lambda} - \mb{e}_{q} \mb{e}_{q}^T -
\mb{I}_n & \mb{0}_n\\\mb{0}_n^T & - 1\end{pmatrix}.
\end{align}
${\bar{\mb{J}}^*}$ is a diagonal matrix, so the eigenvalues can be
obtained directly from the main diagonal. We denote these eigenvalues by
$\alpha_{k}$, $k = 1,\ldots,n+1$, with
\begin{align}
\alpha_{k} & = \lambda_{q}^{-1} \lambda_{k} - 1 = \lambda_{q}^{-1} \mleft( \lambda_{k} - \lambda_{q} \mright)
\quad\mbox{for}\;k=1,\ldots,n,\;k \neq q\\
\alpha_{q} & = \lambda_{q}^{-1} \lambda_{q} - 1 - 1 = - 1 < 0\\
\alpha_{n + 1} & = - 1 < 0.
\end{align}
The first equation shows: Fixed point $q$ is stable (only negative
eigenvalues) if and only if $\lambda_{k} < \lambda_{q}$ for $k =
1\,\ldots,n$, $k \neq q$. Thus the only stable fixed point is found for
$q = 1$ under the assumption that $\lambda_{1}$ is largest eigenvalue of
$\mb{C}$. All other fixed points are saddle points: If $q > 1$, there
exists an index $k$ (particularly $k=1$) for which $\lambda_{k} >
\lambda_{q}$ and thus $\alpha_{k} > 0$. Since the eigenvalues from the
second and third equation are negative, we have a saddle point.

Moreover, for $\lambda_{k} \ll \lambda_{q}$ $\forall k \neq q$ we have
$\alpha_{k} \approx - 1$; we also obtained $-1$ for the remaining
eigenvalues, so approximately the same convergence speed is achieved
from all directions. This demonstrates that the Newton descent solves
the speed-stability problem, at least in the vicinity of the fixed
point.

The Hesse matrix was chosen for the desired ``principal'' fixed point,
and the stability analysis confirms that only this fixed point was
turned into an attractor, thereby substantiating the proposed design
principle.

\subsection{Deflation}\label{sec_deflation}

%
The ordinary differential equation \eqref{eq_newton_principal} is a
single-unit rule which converges towards the principal eigenpair. If
multiple principal eigenpairs have to be determined, we can resort to
deflation of $\mb{C}$. This is based on the removal of preceding
eigenpairs from the spectral decomposition of $\mb{C}$ (introducing zero
eigenvalues). If a learning rule for the principal eigenpair is applied
to the deflated matrix, it will converge to the eigenpair with the
largest eigenvalue remaining in the deflated matrix. If we want to find
the eigenpair $p$, the deflation substitutes $\mb{C}$ with
$\check{\mb{C}}_{p-1}$ where the preceding $p-1$ eigenpairs have been
removed. The removal sets these eigenvalues to zero, such that a
learning rule operating on $\check{\mb{C}}_{p-1}$ converges to the
largest remaining eigenpair, the one which corresponds to the $p$th
largest original eigenvalue. Ultimately, we get a chain of learning
rules which in stage $p$ contains $p - 1$ previous estimator units.

The following derivation of matrix deflation
\citep[][p.39]{nn_Diamantaras96} uses $\mb{C} \mb{v}_{i} = \lambda_{i}
\mb{v}_{i} $ for further transformations, but we only apply the original
equation \eqref{eq_defl1} below:
\begin{align}
&\phantom{{}={}} \check{\mb{C}}_{p-1} \nonumber \\
&\label{eq_defl1}
 = \mb{C} - \sum\limits_{i = 1}^{p - 1}\lambda_{i} \mleft( \mb{v}_{i}
\mb{v}_{i}^T \mright)\\
& = \mb{C} - \sum\limits_{i = 1}^{p - 1}\mb{C} \mleft( \mb{v}_{i}
\mb{v}_{i}^T \mright)\\
& = \mb{C} \mleft( \mb{I}_n - \sum\limits_{i = 1}^{p - 1}\mb{v}_{i}
\mb{v}_{i}^T \mright).
\end{align}
The learning rule for the $p$th eigenpair is obtained by inserting
$\check{\mb{C}}_{p-1}$ from \eqref{eq_defl1} instead of $\mb{C}$ into
\eqref{eq_newton_principal} and applying \eqref{eq_defl1}. We could
perform a fully sequential update scheme where we wait until the
estimate of the first eigenpair has converged to the true eigenpair,
then determine the second eigenpair, and so on. Alternatively, we can
run all estimators in parallel\footnote{The parallel scheme is easier to
implement in a simulation, e.g. in Matlab, and may be faster since later
stages can already start to converge based on halfway converged
estimates from the previous ones. However, the sequential scheme may be
more stable and could allow for larger learning rates. There may be a
good speed-stability compromise between the two schemes.}. For the
parallel scheme, we replace $\mb{w} = \mb{w}_{p}$ and $l = l_{p}$ and
substitute true eigenvectors by estimates $\mb{v}_{i} \approx
\mb{w}_{i}$ and true eigenvalues by estimates $\lambda_{i} \approx
l_{i}$ (these would be the converged eigenpair estimates from the
previous stages in the sequential scheme), and obtain the ODE for the
eigenvector update
\begin{align}
&\phantom{{}={}} \dot{\mb{w}}_{p} \nonumber \\
& = l_{p}^{-1} \mleft( \check{\mb{C}}_{p-1} \mb{w}_{p} - \mleft[
\mb{w}_{p}^T \check{\mb{C}}_{p-1} \mb{w}_{p} \mright] \mb{w}_{p}
\mright) + \frac{1}{2}\, \mleft( \mb{w}_{p}^T \mb{w}_{p} - 1 \mright)
\mb{w}_{p}\\
& = \resizebox{\minof{\width}{\symaeqwidth}}{!}{%
$l_{p}^{-1} \mleft( \mleft[ \mb{C} - \sum\limits_{i = 1}^{p - 1}l_{i}
\mleft\{ \mb{w}_{i} \mb{w}_{i}^T \mright\} \mright] \mb{w}_{p} - \mleft[
\mb{w}_{p}^T \mleft\{ \mb{C} - \sum\limits_{i = 1}^{p - 1}l_{i} \mleft(
\mb{w}_{i} \mb{w}_{i}^T \mright) \mright\} \mb{w}_{p} \mright]
\mb{w}_{p} \mright) + \frac{1}{2}\, \mleft( \mb{w}_{p}^T \mb{w}_{p} - 1
\mright) \mb{w}_{p} $%
}\\
&\label{eq_wdot_defl}
 = \resizebox{\minof{\width}{\symaeqwidth}}{!}{%
$\mleft( \mb{C} \mb{w}_{p} - \sum\limits_{i = 1}^{p - 1}\mb{w}_{i} l_{i}
\mb{w}_{i}^T \mb{w}_{p} - \mleft[ \mb{w}_{p}^T \mb{C} \mb{w}_{p}
\mright] \mb{w}_{p} + \mleft[ \mb{w}_{p}^T \sum\limits_{i = 1}^{p -
1}\mb{w}_{i} l_{i} \mb{w}_{i}^T \mb{w}_{p} \mright] \mb{w}_{p} \mright)
l_{p}^{-1} + \frac{1}{2}\, \mb{w}_{p} \mleft( \mb{w}_{p}^T \mb{w}_{p} -
1 \mright) $%
}.
\end{align}
The same replacement is performed for the eigenvalue update:
\begin{align}
&\phantom{{}={}} \dot{l}_{p} \nonumber \\
& = \mb{w}_{p}^T \check{\mb{C}}_{p-1} \mb{w}_{p} - l_{p} \mb{w}_{p}^T
\mb{w}_{p}\\
& = \mb{w}_{p}^T \mleft( \mb{C} - \sum\limits_{i = 1}^{p - 1}l_{i}
\mb{w}_{i} \mb{w}_{i}^T \mright) \mb{w}_{p} - l_{p} \mb{w}_{p}^T
\mb{w}_{p}\\
&\label{eq_ldot_defl}
 = \mb{w}_{p}^T \mb{C} \mb{w}_{p} - \mb{w}_{p}^T \sum\limits_{i = 1}^{p -
1}\mb{w}_{i} l_{i} \mb{w}_{i}^T \mb{w}_{p} - l_{p} \mb{w}_{p}^T
\mb{w}_{p}.
\end{align}
For the simulation, we switch from vector to matrix form, using Lemma
\ref{lemma_sut_triple} for the treatment of the strict upper triangular
matrices involved. We use $m$ eigenpair estimators. We obtain matrix
equations for eigenvectors
\begin{align}
&\phantom{{}={}} \dot{\mb{W}} \nonumber \\
& = \mleft( \mb{C} \mb{W} - \mb{W} \mb{L} \opnl{sut} \mleft\{ \mb{W}^T
\mb{W} \mright\} \vphantom{\mb{C} \mb{W} - \mb{W} \mb{L} \opnl{sut}
\mleft\{ \mb{W}^T \mb{W} \mright\} - \mb{W} \opnl{diag} \mleft\{
\mb{W}^T \mb{C} \mb{W} \mright\} + \mb{W} \opnl{diag} \mleft\{ \mb{W}^T
\mb{W} \mb{L} \opnl{sut} \mleft\{ \mb{W}^T \mb{W} \mright\} \mright\} }
\mright.
\\&\phantom{{}={}}
\nonumber
{} \! - \mleft. \vphantom{\mb{C} \mb{W} - \mb{W} \mb{L} \opnl{sut}
\mleft\{ \mb{W}^T \mb{W} \mright\} - \mb{W} \opnl{diag} \mleft\{
\mb{W}^T \mb{C} \mb{W} \mright\} + \mb{W} \opnl{diag} \mleft\{ \mb{W}^T
\mb{W} \mb{L} \opnl{sut} \mleft\{ \mb{W}^T \mb{W} \mright\} \mright\} }
\mb{W} \opnl{diag} \mleft\{ \mb{W}^T \mb{C} \mb{W} \mright\} + \mb{W}
\opnl{diag} \mleft\{ \mb{W}^T \mb{W} \mb{L} \opnl{sut} \mleft\{ \mb{W}^T
\mb{W} \mright\} \mright\} \mright) \mb{L}^{-1}
\\&\phantom{{}={}}
\nonumber
{} \! + \frac{1}{2}\, \mb{W} \opnl{diag} \mleft\{ \mb{W}^T \mb{W} -
\mb{I}_m \mright\}
\end{align}
and eigenvalues
\begin{align}
&\phantom{{}={}} \dot{\mb{L}} \nonumber \\
& = \opnl{diag} \mleft\{ \mb{W}^T \mb{C} \mb{W} \mright\} - \opnl{diag}
\mleft\{ \mb{W}^T \mb{W} \mb{L} \opnl{sut} \mleft\{ \mb{W}^T \mb{W}
\mright\} \mright\} - \opnl{diag} \mleft\{ \mb{W}^T \mb{W} \mb{L}
\mright\}.
\end{align}

\section{Arbitrary Eigenpair}\label{sec_arbitrary}

%
We now explore whether we can apply the design principle to derive a
learning rule for an arbitrary eigenpair $p$. We assume that, in each
stage $p$, all previous $p-1$ eigenvalues are known, but not the
following $n-p$ eigenvalues.

\subsection{Inverse Hessian}

%
We start from the transformed Hessian \eqref{eq_Hstar_gen} and
approximate in the vicinity of the desired fixed point $p$ where we have
$l \approx \lambda_{p}$ and $\mb{w} \approx \mb{v}_{p}$:
\begin{align}
&\phantom{{}={}} {\mb{H}^*} \nonumber \\
& \approx \begin{pmatrix}\bs{\Lambda} - l \mb{I}_n & \mb{V}^T \mleft( - \mb{w}
\mright) \\\mleft( - \mb{w}^T \mright) \mb{V} & 0\end{pmatrix}\\
& \approx \begin{pmatrix}\bs{\Lambda} - \lambda_{p} \mb{I}_n & \mb{V}^T \mleft( -
\mb{v}_{p} \mright) \\\mleft( - \mb{v}_{p}^T \mright) \mb{V} &
0\end{pmatrix}\\
&\label{eq_Hstar_p}
 = \begin{pmatrix}\bs{\Lambda} - \lambda_{p} \mb{I}_n & - \mb{e}_{p}\\-
\mb{e}_{p}^T & 0\end{pmatrix}.
\end{align}
The expression in the upper left diagonal matrix of \eqref{eq_Hstar_p}
is treated in the following way: \begin{itemize}

\item We look at the preceding $p-1$ eigenvalues in $\bs{\Lambda}$.
These eigenvalues are much larger than $\lambda_{p}$, i.e. $\lambda_{i}
\gg \lambda_{p}$ for $i < p$. However, we do not introduce
approximations based on this relation, but keep the terms $\lambda_{i} -
\lambda_{p}$.

\item Element $p$ of the upper left diagonal matrix is $0$.

\item We need to eliminate the following $n-p$ eigenvalues since they
are assumed to be unknown for stage $p$. All following $n-p$ eigenvalues
are much smaller than $\lambda_{p}$, i.e. $\lambda_{i} \ll \lambda_{p}$
for $i > p$, and we can therefore approximate $\lambda_{i} - \lambda_{p}
\approx - \lambda_{p}$.

\end{itemize}

In the process of analyzing the case of an arbitrary eigenpair, we
noticed that the fixed points strongly depend on the terms chosen for
the preceding $p-1$ eigenvalues. Undesired stable fixed points were
introduced if approximations were used (tested for constant and linear
Taylor approximation after inversion\footnote{It may be worthwhile to
study whether quadratic approximations eliminate the undesired fixed
points. Approximations might have the advantage that matrix-form
learning rule systems could be derived which is not possible in the
present form.}). In these fixed points, stage $p$ can converge onto the
{\em previous} estimated eigenpair, such that sequences of the same
eigenpair of varying length can appear in the solution (depending on the
initial state). We therefore decided not to use an approximation.

We proceed by approximating $l \approx \lambda_{p}$ and get:
\begin{align}
&\phantom{{}={}} {\mb{H}^*} \nonumber \\
& \approx \begin{pmatrix}\bs{\Lambda}_{p-1} - l \mb{I}_{p-1} & \mb{0}_{p-1} &
\mb{0}_{p-1,n-p} & \mb{0}_{p-1}\\\mb{0}_{p-1}^T & 0 & \mb{0}_{n-p}^T & -
1\\\mb{0}_{p-1,n-p}^T & \mb{0}_{n-p} & - l \mb{I}_{n-p} &
\mb{0}_{n-p}\\\mb{0}_{p-1}^T & - 1 & \mb{0}_{n-p}^T & 0\end{pmatrix}\\
& = \begin{pmatrix}\bs{\Lambda}_{p-1} - l \mb{I}_{p-1} & \mb{0}_{p-1} &
\mb{0}_{p-1,n-p} & \mb{0}_{p-1}\\\mb{0}_{p-1}^T & - 1 & \mb{0}_{n-p}^T &
0\\\mb{0}_{p-1,n-p}^T & \mb{0}_{n-p} & - l \mb{I}_{n-p} &
\mb{0}_{n-p}\\\mb{0}_{p-1}^T & 0 & \mb{0}_{n-p}^T & - 1\end{pmatrix}
\mb{P}_{p,n+1}
\end{align}
where we exchanged column $p$ and $n+1$ in the last step to produce a
diagonal matrix. The inversion below uses $\mb{P}_{p,n+1} =
\mb{P}_{p,n+1}^{-1}$ and performs the corresponding row permutation.

We now aim for a learning rule which rests on the assumption that only
the preceding eigenvectors for $i < p$ are known (either given or
already estimated). This should lead to a learning rule which exhibits
structural similarities to the one derived from deflation in section
\ref{sec_deflation}. In the following transformation, an explicit
dependency on the following eigenvectors for $i > p$ is avoided by
addition of zero on the diagonal and extraction of $l^{-1} \mb{I}_n $.
This leads to a zero block $\mb{0}_{n-p,n-p}$ in the lower right corner;
the zero diagonal elements of this block eliminate the dependency on
$\mb{v}_{i} \mb{v}_{i}^T $ for $i > p$ in the
back-transformation.\footnote{This essentially exploits $\mb{V} \mb{V}^T
= \mb{I}_n$. It may be interesting to explore the derivation of learning
rules where each of the $m \leq n$ eigenpair estimators (index $p \leq
m$) depends on { \em all} $m$ estimated eigenpairs, not just on the
previous ones. The $m$ eigenpairs could be estimated simultaneously.}
\begin{align}
&\phantom{{}={}} {\mb{H}^*}^{-1} \nonumber \\
& \approx \mb{P}_{p,n+1} \begin{pmatrix}\mleft( \bs{\Lambda}_{p-1} - l
\mb{I}_{p-1} \mright)^{-1} & \mb{0}_{p-1} & \mb{0}_{p-1,n-p} &
\mb{0}_{p-1}\\\mb{0}_{p-1}^T & - 1 & \mb{0}_{n-p}^T &
0\\\mb{0}_{p-1,n-p}^T & \mb{0}_{n-p} & - l^{-1} \mb{I}_{n-p} &
\mb{0}_{n-p}\\\mb{0}_{p-1}^T & 0 & \mb{0}_{n-p}^T & - 1\end{pmatrix}\\
& = \begin{pmatrix}\mleft( \bs{\Lambda}_{p-1} - l \mb{I}_{p-1} \mright)^{-1}
& \mb{0}_{p-1} & \mb{0}_{p-1,n-p} & \mb{0}_{p-1}\\\mb{0}_{p-1}^T & 0 &
\mb{0}_{n-p}^T & - 1\\\mb{0}_{p-1,n-p}^T & \mb{0}_{n-p} & - l^{-1}
\mb{I}_{n-p} & \mb{0}_{n-p}\\\mb{0}_{p-1}^T & - 1 & \mb{0}_{n-p}^T &
0\end{pmatrix}\\
& = \begin{pmatrix}\mleft( \bs{\Lambda}_{p-1} - l \mb{I}_{p-1} \mright)^{-1}
- l^{-1} \mb{I}_{p-1} + l^{-1} \mb{I}_{p-1} & \mb{0}_{p-1} &
\mb{0}_{p-1,n-p} & \mb{0}_{p-1}\\\mb{0}_{p-1}^T & \mleft( - l^{-1}
\mright) + l^{-1} & \mb{0}_{n-p}^T & - 1\\\mb{0}_{p-1,n-p}^T &
\mb{0}_{n-p} & - l^{-1} \mb{I}_{n-p} & \mb{0}_{n-p}\\\mb{0}_{p-1}^T & -
1 & \mb{0}_{n-p}^T & 0\end{pmatrix}\\
& = \begin{pmatrix}\begin{pmatrix}\mleft( \bs{\Lambda}_{p-1} - l
\mb{I}_{p-1} \mright)^{-1} + l^{-1} \mb{I}_{p-1} & \mb{0}_{p-1} &
\mb{0}_{p-1,n-p}\\\mb{0}_{p-1}^T & l^{-1} &
\mb{0}_{n-p}^T\\\mb{0}_{p-1,n-p}^T & \mb{0}_{n-p} &
\mb{0}_{n-p,n-p}\end{pmatrix} - l^{-1} \mb{I}_n & - \mb{e}_{p}\\-
\mb{e}_{p}^T & 0\end{pmatrix}.
\end{align}
We transform back and approximate $\mb{w} \approx \mb{v}_{p}$ in the
last step:
\begin{align}
&\phantom{{}={}} \mb{H}^{-1} \nonumber \\
& \approx \mb{T} {\mb{H}^*}^{-1} \mb{T}^T\\
& = \resizebox{\minof{\width}{\symaeqwidth}}{!}{%
$\begin{pmatrix}\mb{V} & \mb{0}_n\\\mb{0}_n^T & 1\end{pmatrix}
\begin{pmatrix}\begin{pmatrix}\mleft( \bs{\Lambda}_{p-1} - l
\mb{I}_{p-1} \mright)^{-1} + l^{-1} \mb{I}_{p-1} & \mb{0}_{p-1} &
\mb{0}_{p-1,n-p}\\\mb{0}_{p-1}^T & l^{-1} &
\mb{0}_{n-p}^T\\\mb{0}_{p-1,n-p}^T & \mb{0}_{n-p} &
\mb{0}_{n-p,n-p}\end{pmatrix} - l^{-1} \mb{I}_n & - \mb{e}_{p}\\-
\mb{e}_{p}^T & 0\end{pmatrix} \begin{pmatrix}\mb{V}^T &
\mb{0}_n\\\mb{0}_n^T & 1\end{pmatrix} $%
}\\
& = \begin{pmatrix}\mb{V} \begin{pmatrix}\mleft( \bs{\Lambda}_{p-1} - l
\mb{I}_{p - 1} \mright)^{-1} + l^{-1} \mb{I}_{p - 1} & \mb{0}_{p - 1} &
\mb{0}_{p - 1,n - p}\\\mb{0}_{p - 1}^T & l^{-1} & \mb{0}_{n -
p}^T\\\mb{0}_{n - p,p - 1} & \mb{0}_{n - p} & \mb{0}_{n - p,n -
p}\end{pmatrix} \mb{V}^T - \mb{V} l^{-1} \mb{V}^T & - \mb{V} \mb{e}_{p}
\\\mleft( - \mb{e}_{p}^T \mright) \mb{V}^T & 0\end{pmatrix}\\
& = \begin{pmatrix}\sum\limits_{i = 1}^{p - 1}\mleft( \mleft[ \lambda_{i} -
l \mright]^{-1} + l^{-1} \mright) \mb{v}_{i} \mb{v}_{i}^T + l^{-1}
\mb{v}_{p} \mb{v}_{p}^T - l^{-1} \mb{I}_n & - \mb{v}_{p}\\- \mb{v}_{p}^T
& 0\end{pmatrix}\\
&\label{eq_Hinv_p}
 \approx \begin{pmatrix}\sum\limits_{i = 1}^{p - 1}\mleft( \mleft[ \lambda_{i} -
l \mright]^{-1} + l^{-1} \mright) \mb{v}_{i} \mb{v}_{i}^T + l^{-1}
\mb{w} \mb{w}^T - l^{-1} \mb{I}_n & - \mb{w}\\- \mb{w}^T &
0\end{pmatrix}.
\end{align}

\subsection{Newton Descent}\label{sec_newton_arbitrary}

%
We insert the approximated inverse Hessian from \eqref{eq_Hinv_p} into
the Newton descent:
\begin{align}
&\phantom{{}={}} \begin{pmatrix}\dot{\mb{w}}\\\dot{l}\end{pmatrix} \nonumber \\
& = - \mb{H}^{-1} \begin{pmatrix}\mleft( \frac{\partial}{\partial \mb{w}}J
\mright)^T\\\mleft( \frac{\partial}{\partial l}J \mright)^T\end{pmatrix}\\
& \approx - \begin{pmatrix}\sum\limits_{i = 1}^{p - 1}\mleft( \mleft[ \lambda_{i}
- l \mright]^{-1} + l^{-1} \mright) \mb{v}_{i} \mb{v}_{i}^T + l^{-1}
\mb{w} \mb{w}^T - l^{-1} \mb{I}_n & - \mb{w}\\- \mb{w}^T &
0\end{pmatrix} \begin{pmatrix}\mb{C} \mb{w} - l \mb{w} \\- \frac{1}{2}\,
\mleft( \mb{w}^T \mb{w} - 1 \mright) \end{pmatrix}\\
& = \resizebox{\minof{\width}{\symaeqwidth}}{!}{%
$- \begin{pmatrix}\sum\limits_{i = 1}^{p - 1}\mleft( \mleft[ \lambda_{i}
- l \mright]^{-1} + l^{-1} \mright) \mb{v}_{i} \mb{v}_{i}^T \mb{C}
\mb{w} - \sum\limits_{i = 1}^{p - 1}\mleft( \mleft[ \lambda_{i} - l
\mright]^{-1} + l^{-1} \mright) \mb{v}_{i} \mb{v}_{i}^T l \mb{w} +
l^{-1} \mb{w} \mb{w}^T \mb{C} \mb{w} - l^{-1} \mb{w} \mb{w}^T l \mb{w} -
l^{-1} \mb{C} \mb{w} + l^{-1} l \mb{w} + \mb{w} \frac{1}{2}\, \mleft(
\mb{w}^T \mb{w} - 1 \mright) \\\mleft( - \mb{w}^T \mb{C} \mb{w} \mright)
+ \mb{w}^T l \mb{w} \end{pmatrix}$%
}\\
& = \resizebox{\minof{\width}{\symaeqwidth}}{!}{%
$\begin{pmatrix}\mleft( - \sum\limits_{i = 1}^{p - 1}\mleft[ \mleft\{
\lambda_{i} - l \mright\}^{-1} + l^{-1} \mright] \mb{v}_{i} \mb{v}_{i}^T
\mb{C} \mb{w} \mright) + l \sum\limits_{i = 1}^{p - 1}\mleft( \mleft[
\lambda_{i} - l \mright]^{-1} + l^{-1} \mright) \mb{v}_{i} \mb{v}_{i}^T
\mb{w} - l^{-1} \mb{w} \mb{w}^T \mb{C} \mb{w} + l^{-1} l \mb{w} \mb{w}^T
\mb{w} + l^{-1} \mb{C} \mb{w} - l^{-1} l \mb{w} - \frac{1}{2}\, \mb{w}
\mb{w}^T \mb{w} + \frac{1}{2}\, \mb{w} \\\mb{w}^T \mb{C} \mb{w} - l
\mb{w}^T \mb{w} \end{pmatrix}$%
}\\
&\label{eq_newton_p_1}
 = \resizebox{\minof{\width}{\symaeqwidth}}{!}{%
$\begin{pmatrix}l^{-1} \mleft( \mb{C} \mb{w} - \mleft[ \mb{w}^T \mb{C}
\mb{w} \mright] \mb{w} \mright) + \frac{1}{2}\, \mb{w} \mleft( \mb{w}^T
\mb{w} - 1 \mright) - \mleft( \sum\limits_{i = 1}^{p - 1}\mleft[
\mleft\{ \lambda_{i} - l \mright\}^{-1} + l^{-1} \mright] \mb{v}_{i}
\mb{v}_{i}^T \mright) \mleft( \mb{C} \mb{w} - l \mb{w} \mright)
\\\mb{w}^T \mb{C} \mb{w} - l \mb{w}^T \mb{w} \end{pmatrix}$%
}.
\end{align}
We see that the first summand of the equation for $\dot{\mb{w}}$
coincides with the learning rule for the principal eigenpair
\eqref{eq_newton_principal}. The second summand depends on the true
previous eigenpairs; its second factor disappears in all fixed points
but the entire summand can affect their stability. It is interesting to
see that the equation for $\dot{l}$ is the same as in
\eqref{eq_newton_principal} (only principal eigenpair estimator),
whereas the deflation procedure for the principal eigenpair leads to an
additional term in equation \eqref{eq_ldot_defl}.

As for the system in section \ref{sec_newton_principal}, we observe
additional fixed points $\mb{w} = \mb{0}_n$ for arbitrary values of $l$
with the exception of the values of $l$ excluded by the inversions.

\subsection{Stability Analysis}

%
In the following, we analyze the stability of the coupled learning rule
\eqref{eq_newton_p_1} in two different ways: by studying the eigenvalues
of the Jacobian (as for the case of the principal eigenpair) and by
studying the effect of small perturbations from the fixed points. The
first way resulted in difficulties for one of the sub-cases since some
terms were undefined (which was not visible any longer after some
transformation steps). The second way coincides with the first for two
of the sub-cases, but allows to study the critical sub-case.

\subsubsection{Stability Analysis via Jacobian}

%
We analyze the stability of the coupled learning rule
\eqref{eq_newton_p_1} by studying the eigenvalues of the Jacobian:
\begin{align}
\begin{pmatrix}\dot{\mb{w}}\\\dot{l}\end{pmatrix}
&=
\resizebox{0.8\width}{!}{%
$\begin{pmatrix}l^{-1} \mleft( \mb{C} \mb{w} - \mleft[ \mb{w}^T \mb{C}
\mb{w} \mright] \mb{w} \mright) + \frac{1}{2}\, \mb{w} \mleft( \mb{w}^T
\mb{w} - 1 \mright) - \mleft( \sum\limits_{i = 1}^{p - 1}\mleft[
\mleft\{ \lambda_{i} - l \mright\}^{-1} + l^{-1} \mright] \mb{v}_{i}
\mb{v}_{i}^T \mright) \mleft( \mb{C} \mb{w} - l \mb{w} \mright)
\\\mb{w}^T \mb{C} \mb{w} - l \mb{w}^T \mb{w} \end{pmatrix}$%
}.
\end{align}
We determine the Jacobian
\begin{align}
&\phantom{{}={}} \mb{J} \nonumber \\
& = \begin{pmatrix}\frac{\partial}{\partial \mb{w}}\dot{\mb{w}} &
\frac{\partial}{\partial l}\dot{\mb{w}}\\\frac{\partial}{\partial
\mb{w}}\dot{l} & \frac{\partial}{\partial l}\dot{l}\end{pmatrix},
\end{align}
the Jacobian at the fixed point
\begin{align}
&\phantom{{}={}} \bar{\mb{J}} \nonumber \\
& = \mleft.\mb{J}\mright|_{\mb{w} = \mb{v}_{q},\; l = \lambda_{q}}\\
& = \begin{pmatrix}\mleft.\mleft( \frac{\partial}{\partial
\mb{w}}\dot{\mb{w}} \mright)\mright|_{\mb{w} = \mb{v}_{q},\; l =
\lambda_{q}} & \mleft.\mleft( \frac{\partial}{\partial l}\dot{\mb{w}}
\mright)\mright|_{\mb{w} = \mb{v}_{q},\; l =
\lambda_{q}}\\\mleft.\mleft( \frac{\partial}{\partial \mb{w}}\dot{l}
\mright)\mright|_{\mb{w} = \mb{v}_{q},\; l = \lambda_{q}} &
\mleft.\mleft( \frac{\partial}{\partial l}\dot{l}
\mright)\mright|_{\mb{w} = \mb{v}_{q},\; l = \lambda_{q}}\end{pmatrix},
\end{align}
and the transformed Jacobian at the fixed point
\begin{align}
&\phantom{{}={}} {\bar{\mb{J}}^*} \nonumber \\
& = \begin{pmatrix}\mb{V} & \mb{0}_n\\\mb{0}_n^T & 1\end{pmatrix}^T
\bar{\mb{J}} \begin{pmatrix}\mb{V} & \mb{0}_n\\\mb{0}_n^T &
1\end{pmatrix}\\
& = \begin{pmatrix}\mb{V} & \mb{0}_n\\\mb{0}_n^T & 1\end{pmatrix}^T
\begin{pmatrix}\mleft.\mleft( \frac{\partial}{\partial
\mb{w}}\dot{\mb{w}} \mright)\mright|_{\mb{w} = \mb{v}_{q},\; l =
\lambda_{q}} & \mleft.\mleft( \frac{\partial}{\partial l}\dot{\mb{w}}
\mright)\mright|_{\mb{w} = \mb{v}_{q},\; l =
\lambda_{q}}\\\mleft.\mleft( \frac{\partial}{\partial \mb{w}}\dot{l}
\mright)\mright|_{\mb{w} = \mb{v}_{q},\; l = \lambda_{q}} &
\mleft.\mleft( \frac{\partial}{\partial l}\dot{l}
\mright)\mright|_{\mb{w} = \mb{v}_{q},\; l = \lambda_{q}}\end{pmatrix}
\begin{pmatrix}\mb{V} & \mb{0}_n\\\mb{0}_n^T & 1\end{pmatrix}\\
& = \begin{pmatrix}\mb{V}^T \mleft.\mleft( \frac{\partial}{\partial
\mb{w}}\dot{\mb{w}} \mright)\mright|_{\mb{w} = \mb{v}_{q},\; l =
\lambda_{q}} \mb{V} & \mb{V}^T \mleft.\mleft( \frac{\partial}{\partial
l}\dot{\mb{w}} \mright)\mright|_{\mb{w} = \mb{v}_{q},\; l = \lambda_{q}}
\\\mleft.\mleft( \frac{\partial}{\partial \mb{w}}\dot{l}
\mright)\mright|_{\mb{w} = \mb{v}_{q},\; l = \lambda_{q}} \mb{V} &
\mleft.\mleft( \frac{\partial}{\partial l}\dot{l}
\mright)\mright|_{\mb{w} = \mb{v}_{q},\; l = \lambda_{q}}\end{pmatrix}.
\end{align}
Since the terms are complex, we proceed for the individual blocks of the
Jacobian. For each block, we first compute the derivative, insert the
fixed point $\begin{pmatrix}\mb{v}_{q}^T & \lambda_{q}\end{pmatrix}^T$
with $\mb{v}_{q}^T \mb{v}_{q} = 1$ and $\mb{C} \mb{v}_{q} = \lambda_{q}
\mb{v}_{q} $, and apply the transformation.

\textbf{Upper left block:}

We repeat the differential equation
\begin{align}
&\phantom{{}={}} \dot{\mb{w}} \nonumber \\
& = l^{-1} \mleft( \mb{C} \mb{w} - \mleft[ \mb{w}^T \mb{C} \mb{w} \mright]
\mb{w} \mright) + \frac{1}{2}\, \mb{w} \mleft( \mb{w}^T \mb{w} - 1
\mright)
\\&\phantom{{}={}}
\nonumber
{} \! - \mleft( \sum\limits_{i = 1}^{p - 1}\mleft[ \mleft\{ \lambda_{i}
- l \mright\}^{-1} + l^{-1} \mright] \mb{v}_{i} \mb{v}_{i}^T \mright)
\mleft( \mb{C} \mb{w} - l \mb{w} \mright),
\end{align}
determine the derivative (using section \ref{app_derivatives})
\begin{align}
&\phantom{{}={}} \frac{\partial}{\partial \mb{w}}\dot{\mb{w}} \nonumber \\
& = l^{-1} \mleft( \mb{C} - 2 \mb{w} \mb{w}^T \mb{C} - \mleft[ \mb{w}^T
\mb{C} \mb{w} \mright] \mb{I}_n \mright) + \mb{w} \mb{w}^T +
\frac{1}{2}\, \mleft( \mb{w}^T \mb{w} \mright) \mb{I}_n - \frac{1}{2}\,
\mb{I}_n
\\&\phantom{{}={}}
\nonumber
{} \! - \mleft( \sum\limits_{i = 1}^{p - 1}\mleft[ \mleft\{ \lambda_{i}
- l \mright\}^{-1} + l^{-1} \mright] \mb{v}_{i} \mb{v}_{i}^T \mright)
\mleft( \mb{C} - l \mb{I}_n \mright),
\end{align}
determine the derivative for the fixed point
\begin{align}
&\phantom{{}={}} \mleft.\mleft( \frac{\partial}{\partial \mb{w}}\dot{\mb{w}}
\mright)\mright|_{\mb{w} = \mb{v}_{q},\; l = \lambda_{q}} \nonumber \\
&\label{eq_problem}
 = \lambda_{q}^{-1} \mleft( \mb{C} - 2 \mb{v}_{q} \mb{v}_{q}^T \mb{C} -
\mleft[ \mb{v}_{q}^T \mb{C} \mb{v}_{q} \mright] \mb{I}_n \mright) +
\mb{v}_{q} \mb{v}_{q}^T + \frac{1}{2}\, \mleft( \mb{v}_{q}^T \mb{v}_{q}
\mright) \mb{I}_n - \frac{1}{2}\, \mb{I}_n
\\&\phantom{{}={}}
\nonumber
{} \! - \mleft( \sum\limits_{i = 1}^{p - 1}\mleft[ \mleft\{ \lambda_{i}
- \lambda_{q} \mright\}^{-1} + \lambda_{q}^{-1} \mright] \mb{v}_{i}
\mb{v}_{i}^T \mright) \mleft( \mb{C} - \lambda_{q} \mb{I}_n \mright)\\
& = \lambda_{q}^{-1} \mleft( \mb{C} - 2 \mb{v}_{q} \mb{v}_{q}^T \lambda_{q}
- \lambda_{q} \mb{I}_n \mright) + \mb{v}_{q} \mb{v}_{q}^T +
\frac{1}{2}\, \mb{I}_n - \frac{1}{2}\, \mb{I}_n
\\&\phantom{{}={}}
\nonumber
{} \! - \sum\limits_{i = 1}^{p - 1}\mleft( \mleft[ \lambda_{i} -
\lambda_{q} \mright]^{-1} + \lambda_{q}^{-1} \mright) \mb{v}_{i}
\mb{v}_{i}^T \mleft( \mb{C} - \lambda_{q} \mb{I}_n \mright)\\
& = \lambda_{q}^{-1} \mb{C} - 2 \mb{v}_{q} \mb{v}_{q}^T - \mb{I}_n +
\mb{v}_{q} \mb{v}_{q}^T - \sum\limits_{i = 1}^{p - 1}\mleft( \mleft[
\lambda_{i} - \lambda_{q} \mright]^{-1} + \lambda_{q}^{-1} \mright)
\mb{v}_{i} \mb{v}_{i}^T \mleft( \lambda_{i} - \lambda_{q} \mright)\\
& = \lambda_{q}^{-1} \mb{C} - \mb{v}_{q} \mb{v}_{q}^T - \mb{I}_n -
\sum\limits_{i = 1}^{p - 1}\mleft( \mleft[ \lambda_{i} - \lambda_{q}
\mright]^{-1} + \lambda_{q}^{-1} \mright) \mb{v}_{i} \mb{v}_{i}^T
\mleft( \lambda_{i} - \lambda_{q} \mright)\\
&\label{eq_problem_disappeared}
 = \lambda_{q}^{-1} \mb{C} - \mb{v}_{q} \mb{v}_{q}^T - \mb{I}_n -
\sum\limits_{i = 1}^{p - 1}\lambda_{q}^{-1} \lambda_{i} \mb{v}_{i}
\mb{v}_{i}^T
\end{align}
(note the $p-1$-fold deflated covariance matrix formed by the first and
last term), and transform:
\begin{align}
&\phantom{{}={}} \mb{V}^T \mleft.\mleft( \frac{\partial}{\partial \mb{w}}\dot{\mb{w}}
\mright)\mright|_{\mb{w} = \mb{v}_{q},\; l = \lambda_{q}} \mb{V} \nonumber \\
& = \mb{V}^T \mleft( \lambda_{q}^{-1} \mb{C} - \mb{v}_{q} \mb{v}_{q}^T -
\mb{I}_n - \sum\limits_{i = 1}^{p - 1}\lambda_{q}^{-1} \lambda_{i}
\mb{v}_{i} \mb{v}_{i}^T \mright) \mb{V}\\
& = \mb{V}^T \mleft( \lambda_{q}^{-1} \mb{C} \mright) \mb{V} - \mb{V}^T
\mleft( \mb{v}_{q} \mb{v}_{q}^T \mright) \mb{V} - \mb{V}^T \mb{I}_n
\mb{V} - \mb{V}^T \sum\limits_{i = 1}^{p - 1}\lambda_{q}^{-1}
\lambda_{i} \mb{v}_{i} \mb{v}_{i}^T \mb{V}\\
& = \lambda_{q}^{-1} \bs{\Lambda} - \mb{e}_{q} \mb{e}_{q}^T - \mb{I}_n -
\sum\limits_{i = 1}^{p - 1}\lambda_{q}^{-1} \lambda_{i} \mb{e}_{i}
\mb{e}_{i}^T\\
& = \lambda_{q}^{-1} \bs{\Lambda} - \mb{e}_{q} \mb{e}_{q}^T - \mb{I}_n -
\lambda_{q}^{-1} \sum\limits_{i = 1}^{p - 1}\lambda_{i} \mb{e}_{i}
\mb{e}_{i}^T\\
& = \lambda_{q}^{-1} \begin{pmatrix}\bs{\Lambda}_{p-1} & \mb{0}_{p-1} &
\mb{0}_{p-1,n-p}\\\mb{0}_{p-1}^T & \lambda_{p} &
\mb{0}_{n-p}^T\\\mb{0}_{p-1,n-p}^T & \mb{0}_{n-p} &
\bs{\Lambda}_{n-p}\end{pmatrix} - \mb{e}_{q} \mb{e}_{q}^T - \mb{I}_n
\\&\phantom{{}={}}
\nonumber
{} \! - \lambda_{q}^{-1} \begin{pmatrix}\bs{\Lambda}_{p-1} &
\mb{0}_{p-1} & \mb{0}_{p-1,n-p}\\\mb{0}_{p-1}^T & 0 &
\mb{0}_{n-p}^T\\\mb{0}_{p-1,n-p}^T & \mb{0}_{n-p} &
\mb{0}_{n-p,n-p}\end{pmatrix}\\
& = \lambda_{q}^{-1} \begin{pmatrix}\mb{0}_{p-1,p-1} & \mb{0}_{p-1} &
\mb{0}_{p-1,n-p}\\\mb{0}_{p-1}^T & \lambda_{p} &
\mb{0}_{n-p}^T\\\mb{0}_{p-1,n-p}^T & \mb{0}_{n-p} &
\bs{\Lambda}_{n-p}\end{pmatrix} - \mb{e}_{q} \mb{e}_{q}^T - \mb{I}_n\\
& = \begin{pmatrix}\mb{0}_{p-1,p-1} & \mb{0}_{p-1} &
\mb{0}_{p-1,n-p}\\\mb{0}_{p-1}^T & \lambda_{q}^{-1} \lambda_{p} &
\mb{0}_{n-p}^T\\\mb{0}_{p-1,n-p}^T & \mb{0}_{n-p} & \lambda_{q}^{-1}
\bs{\Lambda}_{n-p} \end{pmatrix} - \mb{e}_{q} \mb{e}_{q}^T - \mb{I}_n.
\end{align}
\textbf{Upper right block:}

We repeat the differential equation, prepared to reveal dependencies on
$l$
\begin{align}
&\phantom{{}={}} \dot{\mb{w}} \nonumber \\
& = l^{-1} \mleft( \mb{C} \mb{w} - \mleft[ \mb{w}^T \mb{C} \mb{w} \mright]
\mb{w} \mright) + \frac{1}{2}\, \mb{w} \mleft( \mb{w}^T \mb{w} - 1
\mright)
\\&\phantom{{}={}}
\nonumber
{} \! - \mleft( \sum\limits_{i = 1}^{p - 1}\mleft[ \mleft\{ \lambda_{i}
- l \mright\}^{-1} + l^{-1} \mright] \mb{v}_{i} \mb{v}_{i}^T \mright)
\mleft( \mb{C} \mb{w} - l \mb{w} \mright)\\
& = l^{-1} \mleft( \mb{C} \mb{w} - \mleft[ \mb{w}^T \mb{C} \mb{w} \mright]
\mb{w} \mright) + \frac{1}{2}\, \mb{w} \mleft( \mb{w}^T \mb{w} - 1
\mright)
\\&\phantom{{}={}}
\nonumber
{} \! - \sum\limits_{i = 1}^{p - 1}\mleft( \lambda_{i} - l \mright)^{-1}
\mb{v}_{i} \mb{v}_{i}^T \mb{C} \mb{w} - \sum\limits_{i = 1}^{p -
1}l^{-1} \mb{v}_{i} \mb{v}_{i}^T \mb{C} \mb{w} + \sum\limits_{i = 1}^{p
- 1}\mleft( \lambda_{i} - l \mright)^{-1} \mb{v}_{i} \mb{v}_{i}^T l
\mb{w} + \sum\limits_{i = 1}^{p - 1}l^{-1} \mb{v}_{i} \mb{v}_{i}^T l
\mb{w}\\
& = l^{-1} \mleft( \mb{C} \mb{w} - \mleft[ \mb{w}^T \mb{C} \mb{w} \mright]
\mb{w} \mright) + \frac{1}{2}\, \mb{w} \mleft( \mb{w}^T \mb{w} - 1
\mright)
\\&\phantom{{}={}}
\nonumber
{} \! - \sum\limits_{i = 1}^{p - 1}\mleft( \lambda_{i} - l \mright)^{-1}
\mb{v}_{i} \mb{v}_{i}^T \mb{C} \mb{w} - l^{-1} \sum\limits_{i = 1}^{p -
1}\mb{v}_{i} \mb{v}_{i}^T \mb{C} \mb{w} + \sum\limits_{i = 1}^{p -
1}\mleft( \lambda_{i} - l \mright)^{-1} \mb{v}_{i} \mb{v}_{i}^T l \mb{w}
+ \sum\limits_{i = 1}^{p - 1}\mb{v}_{i} \mb{v}_{i}^T \mb{w}\\
& = l^{-1} \mleft( \mb{C} \mb{w} - \mleft[ \mb{w}^T \mb{C} \mb{w} \mright]
\mb{w} \mright) + \frac{1}{2}\, \mb{w} \mleft( \mb{w}^T \mb{w} - 1
\mright)
\\&\phantom{{}={}}
\nonumber
{} \! - \sum\limits_{i = 1}^{p - 1}\mleft( \lambda_{i} - l \mright)^{-1}
\mb{v}_{i} \mb{v}_{i}^T \lambda_{i} \mb{w} - l^{-1} \sum\limits_{i =
1}^{p - 1}\mb{v}_{i} \mb{v}_{i}^T \lambda_{i} \mb{w} + \sum\limits_{i =
1}^{p - 1}\mleft( \lambda_{i} - l \mright)^{-1} \mb{v}_{i} \mb{v}_{i}^T
l \mb{w} + \sum\limits_{i = 1}^{p - 1}\mb{v}_{i} \mb{v}_{i}^T \mb{w}\\
& = l^{-1} \mleft( \mb{C} \mb{w} - \mleft[ \mb{w}^T \mb{C} \mb{w} \mright]
\mb{w} \mright) + \frac{1}{2}\, \mb{w} \mleft( \mb{w}^T \mb{w} - 1
\mright)
\\&\phantom{{}={}}
\nonumber
{} \! - \sum\limits_{i = 1}^{p - 1}\mleft( \lambda_{i} - l \mright)^{-1}
\mb{v}_{i} \mb{v}_{i}^T \mleft( \lambda_{i} - l \mright) \mb{w} - l^{-1}
\sum\limits_{i = 1}^{p - 1}\mb{v}_{i} \mb{v}_{i}^T \lambda_{i} \mb{w} +
\sum\limits_{i = 1}^{p - 1}\mb{v}_{i} \mb{v}_{i}^T \mb{w}\\
& = l^{-1} \mleft( \mb{C} \mb{w} - \mleft[ \mb{w}^T \mb{C} \mb{w} \mright]
\mb{w} \mright) + \frac{1}{2}\, \mb{w} \mleft( \mb{w}^T \mb{w} - 1
\mright)
\\&\phantom{{}={}}
\nonumber
{} \! - \sum\limits_{i = 1}^{p - 1}\mb{v}_{i} \mb{v}_{i}^T \mb{w} -
l^{-1} \sum\limits_{i = 1}^{p - 1}\mb{v}_{i} \mb{v}_{i}^T \lambda_{i}
\mb{w} + \sum\limits_{i = 1}^{p - 1}\mb{v}_{i} \mb{v}_{i}^T \mb{w}\\
& = l^{-1} \mleft( \mb{C} \mb{w} - \mleft[ \mb{w}^T \mb{C} \mb{w} \mright]
\mb{w} \mright) + \frac{1}{2}\, \mb{w} \mleft( \mb{w}^T \mb{w} - 1
\mright) - l^{-1} \sum\limits_{i = 1}^{p - 1}\mb{v}_{i} \mb{v}_{i}^T
\lambda_{i} \mb{w}
\end{align}
(again note the deflation term in the last summand), determine the
derivative
\begin{align}
&\phantom{{}={}} \frac{\partial}{\partial l}\dot{\mb{w}} \nonumber \\
& = \mleft( - l^{-2} \mleft[ \mb{C} \mb{w} - \mleft\{ \mb{w}^T \mb{C} \mb{w}
\mright\} \mb{w} \mright] \mright) + l^{-2} \sum\limits_{i = 1}^{p -
1}\mb{v}_{i} \mb{v}_{i}^T \lambda_{i} \mb{w},
\end{align}
determine the derivative for the fixed point
\begin{align}
&\phantom{{}={}} \mleft.\mleft( \frac{\partial}{\partial l}\dot{\mb{w}}
\mright)\mright|_{\mb{w} = \mb{v}_{q},\; l = \lambda_{q}} \nonumber \\
& = \mleft( - \lambda_{q}^{-2} \mleft[ \mb{C} \mb{v}_{q} - \mleft\{
\mb{v}_{q}^T \mb{C} \mb{v}_{q} \mright\} \mb{v}_{q} \mright] \mright) +
\lambda_{q}^{-2} \sum\limits_{i = 1}^{p - 1}\mb{v}_{i} \mb{v}_{i}^T
\lambda_{i} \mb{v}_{q}\\
& = \mleft( - \lambda_{q}^{-2} \mleft[ \mb{C} \mb{v}_{q} - \lambda_{q}
\mb{v}_{q} \mright] \mright) + \lambda_{q}^{-2} \sum\limits_{i = 1}^{p -
1}\mb{v}_{i} \mb{v}_{i}^T \lambda_{i} \mb{v}_{q}\\
& = \lambda_{q}^{-2} \sum\limits_{i = 1}^{p - 1}\mb{v}_{i} \mb{v}_{i}^T
\lambda_{i} \mb{v}_{q},
\end{align}
and transform:
\begin{align}
&\phantom{{}={}} \mb{V}^T \mleft.\mleft( \frac{\partial}{\partial l}\dot{\mb{w}}
\mright)\mright|_{\mb{w} = \mb{v}_{q},\; l = \lambda_{q}} \nonumber \\
& = \mb{V}^T \mleft( \lambda_{q}^{-2} \sum\limits_{i = 1}^{p - 1}\mb{v}_{i}
\mb{v}_{i}^T \lambda_{i} \mb{v}_{q} \mright)\\
& = \lambda_{q}^{-2} \sum\limits_{i = 1}^{p - 1}\mb{e}_{i} \mb{v}_{i}^T
\lambda_{i} \mb{v}_{q}\\
& = \lambda_{q}^{-2} \sum\limits_{i = 1}^{p - 1}\mb{e}_{i} \lambda_{i}
\delta_{i,q}.
\end{align}
Note that we cannot resolve the sum with the Kronecker delta at this
point since $q$ may be outside the sum's range.

\textbf{Lower left block:}

We repeat the differential equation
\begin{align}
\dot{l}
&=
\mb{w}^T \mb{C} \mb{w} - l \mb{w}^T \mb{w},
\end{align}
determine the derivative
\begin{align}
&\phantom{{}={}} \frac{\partial}{\partial \mb{w}}\dot{l} \nonumber \\
& = 2 \mleft( \mb{w}^T \mb{C} - l \mb{w}^T \mright),
\end{align}
determine the derivative at the fixed point
\begin{align}
&\phantom{{}={}} \mleft.\mleft( \frac{\partial}{\partial \mb{w}}\dot{l}
\mright)\mright|_{\mb{w} = \mb{v}_{q},\; l = \lambda_{q}} \nonumber \\
& = 2 \mleft( \mb{v}_{q}^T \mb{C} - \lambda_{q} \mb{v}_{q}^T \mright)\\
& = \mb{0}_n^T,
\end{align}
and transform:
\begin{align}
&\phantom{{}={}} \mleft.\mleft( \frac{\partial}{\partial \mb{w}}\dot{l}
\mright)\mright|_{\mb{w} = \mb{v}_{q},\; l = \lambda_{q}} \mb{V} \nonumber \\
& = \mb{0}_n^T.
\end{align}
\textbf{Lower right block:}

We repeat the differential equation
\begin{align}
\dot{l}
&=
\mb{w}^T \mb{C} \mb{w} - l \mb{w}^T \mb{w},
\end{align}
determine the derivative
\begin{align}
&\phantom{{}={}} \frac{\partial}{\partial l}\dot{l} \nonumber \\
& = - \mb{w}^T \mb{w},
\end{align}
and determine the derivative for the fixed point
\begin{align}
&\phantom{{}={}} \mleft.\mleft( \frac{\partial}{\partial l}\dot{l}
\mright)\mright|_{\mb{w} = \mb{v}_{q},\; l = \lambda_{q}} \nonumber \\
& = - \mb{v}_{q}^T \mb{v}_{q}\\
& = - 1
\end{align}
(note that the transformation is an identity mapping for lower right
block).

We combine the results for the 4 blocks and obtain
\begin{align}
&\phantom{{}={}} {\bar{\mb{J}}^*} \nonumber \\
& = \begin{pmatrix}\begin{pmatrix}\mb{0}_{p-1,p-1} & \mb{0}_{p-1} &
\mb{0}_{p-1,n-p}\\\mb{0}_{p-1}^T & \lambda_{q}^{-1} \lambda_{p} &
\mb{0}_{n-p}^T\\\mb{0}_{p-1,n-p}^T & \mb{0}_{n-p} & \lambda_{q}^{-1}
\bs{\Lambda}_{n-p} \end{pmatrix} - \mb{e}_{q} \mb{e}_{q}^T - \mb{I}_n &
\lambda_{q}^{-2} \sum\limits_{i = 1}^{p - 1}\mb{e}_{i} \lambda_{i}
\delta_{i,q} \\\mb{0}_n^T & - 1\end{pmatrix}.
\end{align}
Since the analysis of the eigenvalues is complex, we distinguish between
3 different cases.

For the \textbf{first case}, we look at the desired fixed point $q = p$.
The upper right term disappears (since $i \neq q$ for $i =
1,\ldots,p-1$) and we get:
\begin{align}
&\phantom{{}={}} \mleft.{\bar{\mb{J}}^*}\mright|_{q = p} \nonumber \\
& = \begin{pmatrix}\begin{pmatrix}\mb{0}_{p-1,p-1} & \mb{0}_{p-1} &
\mb{0}_{p-1,n-p}\\\mb{0}_{p-1}^T & \lambda_{p}^{-1} \lambda_{p} &
\mb{0}_{n-p}^T\\\mb{0}_{p-1,n-p}^T & \mb{0}_{n-p} & \lambda_{p}^{-1}
\bs{\Lambda}_{n-p} \end{pmatrix} - \mb{e}_{p} \mb{e}_{p}^T - \mb{I}_n &
\mb{0}_n\\\mb{0}_n^T & - 1\end{pmatrix}\\
& = \begin{pmatrix}\begin{pmatrix}\mb{0}_{p-1,p-1} & \mb{0}_{p-1} &
\mb{0}_{p-1,n-p}\\\mb{0}_{p-1}^T & 0 &
\mb{0}_{n-p}^T\\\mb{0}_{p-1,n-p}^T & \mb{0}_{n-p} & \lambda_{p}^{-1}
\bs{\Lambda}_{n-p} \end{pmatrix} - \mb{I}_n & \mb{0}_n\\\mb{0}_n^T & -
1\end{pmatrix}\\
& = \begin{pmatrix}\begin{pmatrix}- \mb{I}_{p-1} & \mb{0}_{p-1} &
\mb{0}_{p-1,n-p}\\\mb{0}_{p-1}^T & - 1 &
\mb{0}_{n-p}^T\\\mb{0}_{p-1,n-p}^T & \mb{0}_{n-p} & \lambda_{p}^{-1}
\bs{\Lambda}_{n-p} - \mb{I}_{n-p}\end{pmatrix} & \mb{0}_n\\\mb{0}_n^T &
- 1\end{pmatrix}.
\end{align}
Since this matrix is diagonal, its eigenvalues $\alpha_{k}$, $k =
1,\ldots,n+1$ appear on the diagonal. We get:
\begin{align}
\alpha_{k} & = - 1 \quad\mbox{for}\;k=1,\ldots,p-1\\
\alpha_{p}
&=
- 1\\
\alpha_{k} & = \lambda_{p}^{-1} \lambda_{k} - 1 \approx - 1 \quad\mbox{for}\;k=p+1,\ldots,n,\; \lambda_{k} \ll \lambda_{p}\\
\alpha_{n + 1}
&=
- 1.
\end{align}
We see that fixed point $p$ is an attractor, and the convergence speed
is about the same from all directions.

Note that stability only requires relations such as $\lambda_{k} <
\lambda_{p}$ to ensure negative eigenvalues of the Jacobian, but not
$\lambda_{k} \ll \lambda_{p}$ which just guarantees that the eigenvalues
are close to $-1$. However, the convergence speed depends on the
differences between the eigenvalues of $\mb{C}$.

For the \textbf{second case}, we look at fixed point $q$ which
corresponds to an eigenpair {\em following} $p$, i.e. $q > p$. Again,
the upper right term disappears and we get:
\begin{align}
&\phantom{{}={}} \mleft.{\bar{\mb{J}}^*}\mright|_{q > p} \nonumber \\
& = \begin{pmatrix}\begin{pmatrix}\mb{0}_{p-1,p-1} & \mb{0}_{p-1} &
\mb{0}_{p-1,n-p}\\\mb{0}_{p-1}^T & \lambda_{q}^{-1} \lambda_{p} &
\mb{0}_{n-p}^T\\\mb{0}_{p-1,n-p}^T & \mb{0}_{n-p} & \lambda_{q}^{-1}
\bs{\Lambda}_{n-p} \end{pmatrix} - \mb{e}_{q} \mb{e}_{q}^T - \mb{I}_n &
\lambda_{q}^{-2} \sum\limits_{i = 1}^{p - 1}\mb{e}_{i} \lambda_{i}
\delta_{i,q} \\\mb{0}_n^T & - 1\end{pmatrix}\\
& = \begin{pmatrix}\begin{pmatrix}\mb{0}_{p-1,p-1} & \mb{0}_{p-1} &
\mb{0}_{p-1,n-p}\\\mb{0}_{p-1}^T & \lambda_{q}^{-1} \lambda_{p} &
\mb{0}_{n-p}^T\\\mb{0}_{p-1,n-p}^T & \mb{0}_{n-p} & \lambda_{q}^{-1}
\bs{\Lambda}_{n-p} \end{pmatrix} - \mb{e}_{q} \mb{e}_{q}^T - \mb{I}_n &
\mb{0}_n\\\mb{0}_n^T & - 1\end{pmatrix}\\
& = \begin{pmatrix}\begin{pmatrix}- \mb{I}_{p-1} & \mb{0}_{p-1} &
\mb{0}_{p-1,n-p}\\\mb{0}_{p-1}^T & \lambda_{q}^{-1} \lambda_{p} - 1 &
\mb{0}_{n-p}^T\\\mb{0}_{p-1,n-p}^T & \mb{0}_{n-p} & \lambda_{q}^{-1}
\bs{\Lambda}_{n-p} - \mb{I}_{n-p}\end{pmatrix} - \mb{e}_{q} \mb{e}_{q}^T
& \mb{0}_n\\\mb{0}_n^T & - 1\end{pmatrix}
\end{align}
The eigenvalues $\alpha_{k}$, $k = 1,\ldots,n+1$, can again be read out
from the diagonal. Index $q$ lies in the lower right block of the inner
block matrix, so we get eigenvalues for the upper left block, for the
middle block, and for 3 different regions in the lower right block
(before, at, and after $q$):
\begin{align}
\alpha_{k} & = - 1 < 0 \quad\mbox{for}\;k=1,\ldots,p-1\\
\alpha_{p} & = \lambda_{q}^{-1} \lambda_{p} - 1 > 0 \quad\mbox{for}\;p < q,\;\lambda_{p} \gg \lambda_{q}\\
\alpha_{k} & = \lambda_{q}^{-1} \lambda_{k} - 1 > 0 \quad\mbox{for}\;k=p+1,\ldots,q-1,\; \lambda_{k} \gg \lambda_{q}\\
\alpha_{q} & = \lambda_{q}^{-1} \lambda_{q} - 2 = - 1 < 0\\
\alpha_{k} & = \lambda_{q}^{-1} \lambda_{k} - 1 < 0 \quad\mbox{for}\;k=q+1,\ldots,n,\; \lambda_{k} \ll \lambda_{q}\\
\alpha_{n + 1} & = - 1 < 0
\end{align}
We conclude that the fixed points following $p$ are always saddle
points.

The \textbf{third case} is the critical case and concerns $q < p$, i.e.
fixed points preceding $p$. Here the upper right term does not
disappear. We get
\begin{align}
&\phantom{{}={}} \mleft.{\bar{\mb{J}}^*}\mright|_{q < p} \nonumber \\
& = \begin{pmatrix}\begin{pmatrix}\mb{0}_{p-1,p-1} & \mb{0}_{p-1} &
\mb{0}_{p-1,n-p}\\\mb{0}_{p-1}^T & \lambda_{q}^{-1} \lambda_{p} &
\mb{0}_{n-p}^T\\\mb{0}_{p-1,n-p}^T & \mb{0}_{n-p} & \lambda_{q}^{-1}
\bs{\Lambda}_{n-p} \end{pmatrix} - \mb{e}_{q} \mb{e}_{q}^T - \mb{I}_n &
\lambda_{q}^{-2} \sum\limits_{i = 1}^{p - 1}\mb{e}_{i} \lambda_{i}
\delta_{i,q} \\\mb{0}_n^T & - 1\end{pmatrix}\\
& = \begin{pmatrix}\begin{pmatrix}\mb{0}_{p-1,p-1} & \mb{0}_{p-1} &
\mb{0}_{p-1,n-p}\\\mb{0}_{p-1}^T & \lambda_{q}^{-1} \lambda_{p} &
\mb{0}_{n-p}^T\\\mb{0}_{p-1,n-p}^T & \mb{0}_{n-p} & \lambda_{q}^{-1}
\bs{\Lambda}_{n-p} \end{pmatrix} - \mb{e}_{q} \mb{e}_{q}^T - \mb{I}_n &
\lambda_{q}^{-1} \mb{e}_{q} \\\mb{0}_n^T & - 1\end{pmatrix}\\
& = \begin{pmatrix}\begin{pmatrix}- \mb{I}_{p-1} & \mb{0}_{p-1} &
\mb{0}_{p-1,n-p}\\\mb{0}_{p-1}^T & \lambda_{q}^{-1} \lambda_{p} - 1 &
\mb{0}_{n-p}^T\\\mb{0}_{p-1,n-p}^T & \mb{0}_{n-p} & \lambda_{q}^{-1}
\bs{\Lambda}_{n-p} - \mb{I}_{n-p}\end{pmatrix} - \mb{e}_{q} \mb{e}_{q}^T
& \lambda_{q}^{-1} \mb{e}_{q} \\\mb{0}_n^T & - 1\end{pmatrix}
\end{align}
The single off-diagonal element at row $q$ of column $n+1$ doesn't
affect eigenvalues, since the matrix is triangular and still all
eigenvalues $\alpha_{k}$, $k = 1,\ldots,n+1$ are found on the main
diagonal. Index $q$ lies in the upper left block, so we get eigenvalues
for 3 different regions in the upper left block (before, at, and after
$q$), for the middle block, and for the lower right block:
\begin{align}
\alpha_{k} & = - 1 < 0 \quad\mbox{for}\;k=1,\ldots,q-1\\
\alpha_{q} & = - 2 < 0\\
\alpha_{k} & = - 1 < 0 \quad\mbox{for}\;k=q+1,\ldots,p-1\\
\alpha_{p} & = \lambda_{q}^{-1} \lambda_{p} - 1 < 0 \quad\mbox{for}\;q < p,\; \lambda_{q} \gg \lambda_{p}\\
\alpha_{k} & = \lambda_{q}^{-1} \lambda_{k} - 1 < 0 \quad\mbox{for}\;k=p+1,\ldots,n,\; \lambda_{k} \ll \lambda_{q}\\
\alpha_{n + 1} & = - 1 < 0.
\end{align}
We would have to conclude from this result that there are additional
stable fixed points at undesired indices $q < p$. However, this stands
in conflict with observations from simulations which prompted us to
re-examine the stability analysis. The problem is visible in equation
\eqref{eq_problem}: For $q < p$, one term $\mleft( \lambda_{i} -
\lambda_{q} \mright)^{-1}$ for summation indices $i = 1 \ldots p-1$ is
undefined. The other two stability cases are unaffected, since for $q >
p$, the index $q$ lies outside the summation range. Unfortunately, the
problem is no longer visible in \eqref{eq_problem_disappeared} since the
critical terms have disappeared. Nevertheless, the analysis of the third
case above is flawed since the fixed points under consideration don't
even exist any longer after the multiplication with the inverse Hessian
in the Newton descent.

\subsubsection{Stability Analysis via Perturbation}

%
To avoid the problem with the stability analysis via the eigenvalues of
the Jacobian, we re-examine stability by a different method: We
introduce small perturbations at the original fixed points of the
Lagrange criterion (i.e. including those that become undefined after the
inversion of the Hessian in the Newton descent). At the point of the
perturbed state, we analyze the scalar product between the perturbation
and the direction of movement. If we can show that this scalar product
is negative for all perturbations, the point under consideration is
approached. If there are perturbations where the scalar product is
positive, there are directions where the movement leads away from this
point. Since the perturbations are small, we can eliminate all terms of
higher order, if terms of lower order exist.

We start from \eqref{eq_newton_p_1}. A perturbation adds small changes
to an original fixed point $q$, i.e. we set $\mb{w} = \mb{v}_{q} +
\bs{\mu}$ and $l = \lambda_{q} + \nu$:
\begin{align}
&\phantom{{}={}} \dot{\mb{w}} \nonumber \\
& = l^{-1} \mleft( \mb{C} \mb{w} - \mleft[ \mb{w}^T \mb{C} \mb{w} \mright]
\mb{w} \mright)
\\&\phantom{{}={}}
\nonumber
{} \! + \frac{1}{2}\, \mb{w} \mleft( \mb{w}^T \mb{w} - 1 \mright)
\\&\phantom{{}={}}
\nonumber
{} \! - \mleft( \sum\limits_{i = 1}^{p - 1}\mleft[ \mleft\{ \lambda_{i}
- l \mright\}^{-1} + l^{-1} \mright] \mb{v}_{i} \mb{v}_{i}^T \mright)
\mleft( \mb{C} \mb{w} - l \mb{w} \mright)\\
& = \mleft( \lambda_{q} + \nu \mright)^{-1} \mleft( \mb{C} \mleft[
\mb{v}_{q} + \bs{\mu} \mright] - \mleft[ \mleft\{ \mb{v}_{q} + \bs{\mu}
\mright\}^T \mb{C} \mleft\{ \mb{v}_{q} + \bs{\mu} \mright\} \mright]
\mleft[ \mb{v}_{q} + \bs{\mu} \mright] \mright)
\\&\phantom{{}={}}
\nonumber
{} \! + \frac{1}{2}\, \mleft( \mb{v}_{q} + \bs{\mu} \mright) \mleft(
\mleft[ \mb{v}_{q} + \bs{\mu} \mright]^T \mleft[ \mb{v}_{q} + \bs{\mu}
\mright] - 1 \mright)
\\&\phantom{{}={}}
\nonumber
{} \! - \mleft( \sum\limits_{i = 1}^{p - 1}\mleft[ \mleft\{ \lambda_{i}
- \mleft( \lambda_{q} + \nu \mright) \mright\}^{-1} + \mleft\{
\lambda_{q} + \nu \mright\}^{-1} \mright] \mb{v}_{i} \mb{v}_{i}^T
\mright) \mleft( \mb{C} \mleft[ \mb{v}_{q} + \bs{\mu} \mright] - \mleft[
\lambda_{q} + \nu \mright] \mleft[ \mb{v}_{q} + \bs{\mu} \mright]
\mright)\\
& = \resizebox{0.6\width}{!}{%
$\mleft( \lambda_{q} + \nu \mright)^{-1} \mleft( \mb{C} \mb{v}_{q} +
\mb{C} \bs{\mu} - \mleft[ \mb{v}_{q}^T \mb{C} \mb{v}_{q} \mright]
\mb{v}_{q} - \mleft[ \bs{\mu}^T \mb{C} \mb{v}_{q} \mright] \mb{v}_{q} -
\mleft[ \mb{v}_{q}^T \mb{C} \bs{\mu} \mright] \mb{v}_{q} - \mleft[
\bs{\mu}^T \mb{C} \bs{\mu} \mright] \mb{v}_{q} - \mleft[ \mb{v}_{q}^T
\mb{C} \mb{v}_{q} \mright] \bs{\mu} - \mleft[ \bs{\mu}^T \mb{C}
\mb{v}_{q} \mright] \bs{\mu} - \mleft[ \mb{v}_{q}^T \mb{C} \bs{\mu}
\mright] \bs{\mu} - \mleft[ \bs{\mu}^T \mb{C} \bs{\mu} \mright] \bs{\mu}
\mright) $%
}
\\&\phantom{{}={}}
\nonumber
{} \! + \resizebox{0.8\width}{!}{%
$\frac{1}{2}\, \mleft( \mb{v}_{q} \mb{v}_{q}^T \mb{v}_{q} + \mb{v}_{q}
\mb{v}_{q}^T \bs{\mu} + \mb{v}_{q} \bs{\mu}^T \mb{v}_{q} + \mb{v}_{q}
\bs{\mu}^T \bs{\mu} - \mb{v}_{q} + \bs{\mu} \mb{v}_{q}^T \mb{v}_{q} +
\bs{\mu} \mb{v}_{q}^T \bs{\mu} + \bs{\mu} \bs{\mu}^T \mb{v}_{q} +
\bs{\mu} \bs{\mu}^T \bs{\mu} - \bs{\mu} \mright) $%
}
\\&\phantom{{}={}}
\nonumber
{} \! - \resizebox{0.8\width}{!}{%
$\mleft( \sum\limits_{i = 1}^{p - 1}\mleft[ \mleft\{ \lambda_{i} -
\mleft( \lambda_{q} + \nu \mright) \mright\}^{-1} + \mleft\{ \lambda_{q}
+ \nu \mright\}^{-1} \mright] \mb{v}_{i} \mb{v}_{i}^T \mright) \mleft(
\mb{C} \mb{v}_{q} + \mb{C} \bs{\mu} - \lambda_{q} \mb{v}_{q} -
\lambda_{q} \bs{\mu} - \nu \mb{v}_{q} - \nu \bs{\mu} \mright) $%
}\\
& \approx \mleft( \lambda_{q} + \nu \mright)^{-1} \mleft( \mb{C} \bs{\mu} - 2
\mleft[ \bs{\mu}^T \lambda_{q} \mb{v}_{q} \mright] \mb{v}_{q} -
\lambda_{q} \bs{\mu} \mright)
\\&\phantom{{}={}}
\nonumber
{} \! + \mb{v}_{q} \mb{v}_{q}^T \bs{\mu}
\\&\phantom{{}={}}
\nonumber
{} \! - \mleft( \sum\limits_{i = 1}^{p - 1}\mleft[ \mleft\{ \lambda_{i}
- \mleft( \lambda_{q} + \nu \mright) \mright\}^{-1} + \mleft\{
\lambda_{q} + \nu \mright\}^{-1} \mright] \mb{v}_{i} \mb{v}_{i}^T
\mright) \mleft( \mb{C} \bs{\mu} - \lambda_{q} \bs{\mu} - \nu \mb{v}_{q}
\mright)\\
& \approx \lambda_{q}^{-1} \mleft( \mb{C} \bs{\mu} - 2 \mleft[ \bs{\mu}^T
\lambda_{q} \mb{v}_{q} \mright] \mb{v}_{q} - \lambda_{q} \bs{\mu}
\mright)
\\&\phantom{{}={}}
\nonumber
{} \! + \mb{v}_{q} \mb{v}_{q}^T \bs{\mu}
\\&\phantom{{}={}}
\nonumber
{} \! - \mleft( \sum\limits_{i = 1}^{p - 1}\mleft[ \mleft\{ \lambda_{i}
- \mleft( \lambda_{q} + \nu \mright) \mright\}^{-1} + \lambda_{q}^{-1}
\mright] \mb{v}_{i} \mb{v}_{i}^T \mright) \mleft( \mb{C} \bs{\mu} -
\lambda_{q} \bs{\mu} - \nu \mb{v}_{q} \mright)\\
& = \lambda_{q}^{-1} \mb{C} \bs{\mu} - \mb{v}_{q} \mb{v}_{q}^T \bs{\mu} -
\bs{\mu}
\\&\phantom{{}={}}
\nonumber
{} \! - \sum\limits_{i = 1}^{p - 1}\mleft( \mleft[ \lambda_{i} -
\mleft\{ \lambda_{q} + \nu \mright\} \mright]^{-1} + \lambda_{q}^{-1}
\mright) \mb{v}_{i} \mb{v}_{i}^T \mleft( \lambda_{i} \bs{\mu} -
\lambda_{q} \bs{\mu} - \nu \mb{v}_{q} \mright)
\end{align}
where we approximated $\mleft( \lambda_{q} + \nu \mright)^{-1} \approx
\lambda_{q}^{-1} - \nu \lambda_{q}^{-2} $ and omitted terms of second
order and above.

The same perturbation transformation is done for $\dot{l}$:
\begin{align}
&\phantom{{}={}} \dot{l} \nonumber \\
& = \mb{w}^T \mb{C} \mb{w} - l \mb{w}^T \mb{w}\\
& = \mleft( \mb{v}_{q} + \bs{\mu} \mright)^T \mb{C} \mleft( \mb{v}_{q} +
\bs{\mu} \mright) - \mleft( \lambda_{q} + \nu \mright) \mleft(
\mb{v}_{q} + \bs{\mu} \mright)^T \mleft( \mb{v}_{q} + \bs{\mu} \mright)\\
& = \mb{v}_{q}^T \mb{C} \mb{v}_{q} + \bs{\mu}^T \mb{C} \mb{v}_{q} +
\mb{v}_{q}^T \mb{C} \bs{\mu} + \bs{\mu}^T \mb{C} \bs{\mu}
\\&\phantom{{}={}}
\nonumber
{} \! - \lambda_{q} \mb{v}_{q}^T \mb{v}_{q} - \lambda_{q} \bs{\mu}^T
\mb{v}_{q} - \lambda_{q} \mb{v}_{q}^T \bs{\mu} - \lambda_{q} \bs{\mu}^T
\bs{\mu} - \nu \mb{v}_{q}^T \mb{v}_{q} - \nu \bs{\mu}^T \mb{v}_{q} - \nu
\mb{v}_{q}^T \bs{\mu} - \nu \bs{\mu}^T \bs{\mu}\\
& \approx - \nu.
\end{align}
We now determine the scalar product between perturbation and movement
direction:
\begin{align}
&\phantom{{}={}} \begin{pmatrix}\bs{\mu}\\\nu\end{pmatrix}^T
\begin{pmatrix}\dot{\mb{w}}\\\dot{l}\end{pmatrix} \nonumber \\
& = \bs{\mu}^T \dot{\mb{w}} + \nu \dot{l}\\
& \approx \lambda_{q}^{-1} \bs{\mu}^T \mb{C} \bs{\mu} - \bs{\mu}^T \mb{v}_{q}
\mb{v}_{q}^T \bs{\mu} - \bs{\mu}^T \bs{\mu} - \nu^2
\\&\phantom{{}={}}
\nonumber
{} \! - \sum\limits_{i = 1}^{p - 1}\mleft( \mleft[ \lambda_{i} -
\mleft\{ \lambda_{q} + \nu \mright\} \mright]^{-1} + \lambda_{q}^{-1}
\mright) \bs{\mu}^T \mb{v}_{i} \mb{v}_{i}^T \mleft( \lambda_{i} \bs{\mu}
- \lambda_{q} \bs{\mu} - \nu \mb{v}_{q} \mright)\\
& = \lambda_{q}^{-1} \bs{\mu}^T \mb{C} \bs{\mu} - \mleft( \mb{v}_{q}^T
\bs{\mu} \mright)^2 - \bs{\mu}^T \bs{\mu} - \nu^2
\\&\phantom{{}={}}
\nonumber
{} \! - \sum\limits_{i = 1}^{p - 1}\mleft( \mleft[ \lambda_{i} -
\mleft\{ \lambda_{q} + \nu \mright\} \mright]^{-1} + \lambda_{q}^{-1}
\mright) \mleft( \lambda_{i} - \lambda_{q} \mright) \mleft( \mb{v}_{i}^T
\bs{\mu} \mright)^2
\\&\phantom{{}={}}
\nonumber
{} \! + \sum\limits_{i = 1}^{p - 1}\mleft( \mleft[ \lambda_{i} -
\mleft\{ \lambda_{q} + \nu \mright\} \mright]^{-1} + \lambda_{q}^{-1}
\mright) \nu \bs{\mu}^T \mb{v}_{i} \mb{v}_{i}^T \mb{v}_{q}\\
& = \lambda_{q}^{-1} \sum\limits_{i = 1}^{n}\bs{\mu}^T \mb{v}_{i}
\lambda_{i} \mb{v}_{i}^T \bs{\mu} - \mleft( \mb{v}_{q}^T \bs{\mu}
\mright)^2 - \bs{\mu}^T \bs{\mu} - \nu^2
\\&\phantom{{}={}}
\nonumber
{} \! - \sum\limits_{i = 1}^{p - 1}\mleft( \mleft[ \lambda_{i} -
\mleft\{ \lambda_{q} + \nu \mright\} \mright]^{-1} + \lambda_{q}^{-1}
\mright) \mleft( \lambda_{i} - \lambda_{q} \mright) \mleft( \mb{v}_{i}^T
\bs{\mu} \mright)^2
\\&\phantom{{}={}}
\nonumber
{} \! + \sum\limits_{i = 1}^{p - 1}\mleft( \mleft[ \lambda_{i} -
\mleft\{ \lambda_{q} + \nu \mright\} \mright]^{-1} + \lambda_{q}^{-1}
\mright) \nu \bs{\mu}^T \mb{v}_{i} \delta_{i,q}\\
& = \sum\limits_{i = 1}^{n}\lambda_{q}^{-1} \lambda_{i} \mleft( \mb{v}_{i}^T
\bs{\mu} \mright)^2 - \mleft( \mb{v}_{q}^T \bs{\mu} \mright)^2 -
\bs{\mu}^T \bs{\mu} - \nu^2
\\&\phantom{{}={}}
\nonumber
{} \! - \sum\limits_{i = 1}^{p - 1}\mleft( \mleft[ \lambda_{i} -
\mleft\{ \lambda_{q} + \nu \mright\} \mright]^{-1} + \lambda_{q}^{-1}
\mright) \mleft( \lambda_{i} - \lambda_{q} \mright) \mleft( \mb{v}_{i}^T
\bs{\mu} \mright)^2
\\&\phantom{{}={}}
\nonumber
{} \! + \sum\limits_{i = 1}^{p - 1}\mleft( \mleft[ \lambda_{i} -
\mleft\{ \lambda_{q} + \nu \mright\} \mright]^{-1} + \lambda_{q}^{-1}
\mright) \nu \bs{\mu}^T \mb{v}_{i} \delta_{i,q}\\
& = \sum\limits_{i = 1}^{n}\lambda_{q}^{-1} \lambda_{i} \mleft( \mb{v}_{i}^T
\bs{\mu} \mright)^2 - \mleft( \mb{v}_{q}^T \bs{\mu} \mright)^2 -
\bs{\mu}^T \sum\limits_{i = 1}^{n}\mb{v}_{i} \mb{v}_{i}^T \bs{\mu} -
\nu^2
\\&\phantom{{}={}}
\nonumber
{} \! - \sum\limits_{i = 1}^{p - 1}\mleft( \mleft[ \lambda_{i} -
\mleft\{ \lambda_{q} + \nu \mright\} \mright]^{-1} + \lambda_{q}^{-1}
\mright) \mleft( \lambda_{i} - \lambda_{q} \mright) \mleft( \mb{v}_{i}^T
\bs{\mu} \mright)^2
\\&\phantom{{}={}}
\nonumber
{} \! + \sum\limits_{i = 1}^{p - 1}\mleft( \mleft[ \lambda_{i} -
\mleft\{ \lambda_{q} + \nu \mright\} \mright]^{-1} + \lambda_{q}^{-1}
\mright) \nu \bs{\mu}^T \mb{v}_{i} \delta_{i,q}\\
& = \sum\limits_{i = 1}^{n}\lambda_{q}^{-1} \lambda_{i} \mleft( \mb{v}_{i}^T
\bs{\mu} \mright)^2 - \mleft( \mb{v}_{q}^T \bs{\mu} \mright)^2 -
\sum\limits_{i = 1}^{n}\mleft( \mb{v}_{i}^T \bs{\mu} \mright)^2 - \nu^2
\\&\phantom{{}={}}
\nonumber
{} \! - \sum\limits_{i = 1}^{p - 1}\mleft( \mleft[ \lambda_{i} -
\mleft\{ \lambda_{q} + \nu \mright\} \mright]^{-1} + \lambda_{q}^{-1}
\mright) \mleft( \lambda_{i} - \lambda_{q} \mright) \mleft( \mb{v}_{i}^T
\bs{\mu} \mright)^2
\\&\phantom{{}={}}
\nonumber
{} \! + \sum\limits_{i = 1}^{p - 1}\mleft( \mleft[ \lambda_{i} -
\mleft\{ \lambda_{q} + \nu \mright\} \mright]^{-1} + \lambda_{q}^{-1}
\mright) \nu \bs{\mu}^T \mb{v}_{i} \delta_{i,q}\\
& = \sum\limits_{i = 1}^{n}\mleft( \lambda_{q}^{-1} \lambda_{i} - 1 \mright)
\mleft( \mb{v}_{i}^T \bs{\mu} \mright)^2 - \mleft( \mb{v}_{q}^T \bs{\mu}
\mright)^2 - \nu^2
\\&\phantom{{}={}}
\nonumber
{} \! - \sum\limits_{i = 1}^{p - 1}\mleft( \mleft[ \lambda_{i} -
\mleft\{ \lambda_{q} + \nu \mright\} \mright]^{-1} + \lambda_{q}^{-1}
\mright) \mleft( \lambda_{i} - \lambda_{q} \mright) \mleft( \mb{v}_{i}^T
\bs{\mu} \mright)^2
\\&\phantom{{}={}}
\nonumber
{} \! + \sum\limits_{i = 1}^{p - 1}\mleft( \mleft[ \lambda_{i} -
\mleft\{ \lambda_{q} + \nu \mright\} \mright]^{-1} + \lambda_{q}^{-1}
\mright) \nu \bs{\mu}^T \mb{v}_{i} \delta_{i,q}.
\end{align}
We study all three stability cases as before, even though only one is
critical; we use the non-critical cases to check whether the results
coincide with the former analysis.

We start with the \textbf{first case} $q = p$:
\begin{align}
&\phantom{{}={}} \sum\limits_{i = 1}^{n}\mleft( \lambda_{q}^{-1} \lambda_{i} - 1 \mright)
\mleft( \mb{v}_{i}^T \bs{\mu} \mright)^2 - \mleft( \mb{v}_{q}^T \bs{\mu}
\mright)^2 - \nu^2
\\&\phantom{{}={}}
\nonumber
{} \! - \sum\limits_{i = 1}^{p - 1}\mleft( \mleft[ \lambda_{i} -
\mleft\{ \lambda_{q} + \nu \mright\} \mright]^{-1} + \lambda_{q}^{-1}
\mright) \mleft( \lambda_{i} - \lambda_{q} \mright) \mleft( \mb{v}_{i}^T
\bs{\mu} \mright)^2
\\&\phantom{{}={}}
\nonumber
{} \! + \sum\limits_{i = 1}^{p - 1}\mleft( \mleft[ \lambda_{i} -
\mleft\{ \lambda_{q} + \nu \mright\} \mright]^{-1} + \lambda_{q}^{-1}
\mright) \nu \bs{\mu}^T \mb{v}_{i} \delta_{i,q} \nonumber \\
& = \sum\limits_{i = 1}^{n}\mleft( \lambda_{p}^{-1} \lambda_{i} - 1 \mright)
\mleft( \mb{v}_{i}^T \bs{\mu} \mright)^2 - \mleft( \mb{v}_{p}^T \bs{\mu}
\mright)^2 - \nu^2
\\&\phantom{{}={}}
\nonumber
{} \! - \sum\limits_{i = 1}^{p - 1}\mleft( \mleft[ \lambda_{i} -
\mleft\{ \lambda_{p} + \nu \mright\} \mright]^{-1} + \lambda_{p}^{-1}
\mright) \mleft( \lambda_{i} - \lambda_{p} \mright) \mleft( \mb{v}_{i}^T
\bs{\mu} \mright)^2\\
& \approx \sum\limits_{i = 1}^{n}\mleft( \lambda_{p}^{-1} \lambda_{i} - 1 \mright)
\mleft( \mb{v}_{i}^T \bs{\mu} \mright)^2 - \mleft( \mb{v}_{p}^T \bs{\mu}
\mright)^2 - \nu^2
\\&\phantom{{}={}}
\nonumber
{} \! - \sum\limits_{i = 1}^{p - 1}\mleft( \mleft[ \lambda_{i} -
\lambda_{p} \mright]^{-1} + \lambda_{p}^{-1} \mright) \mleft(
\lambda_{i} - \lambda_{p} \mright) \mleft( \mb{v}_{i}^T \bs{\mu}
\mright)^2\\
& = \sum\limits_{i = 1}^{n}\mleft( \lambda_{p}^{-1} \lambda_{i} - 1 \mright)
\mleft( \mb{v}_{i}^T \bs{\mu} \mright)^2 - \mleft( \mb{v}_{p}^T \bs{\mu}
\mright)^2 - \nu^2 - \sum\limits_{i = 1}^{p - 1}\lambda_{p}^{-1}
\lambda_{i} \mleft( \mb{v}_{i}^T \bs{\mu} \mright)^2\\
& = \sum\limits_{i = 1}^{n}\lambda_{p}^{-1} \lambda_{i} \mleft( \mb{v}_{i}^T
\bs{\mu} \mright)^2 - \sum\limits_{i = 1}^{n}\mleft( \mb{v}_{i}^T
\bs{\mu} \mright)^2 - \mleft( \mb{v}_{p}^T \bs{\mu} \mright)^2 - \nu^2 -
\sum\limits_{i = 1}^{p - 1}\lambda_{p}^{-1} \lambda_{i} \mleft(
\mb{v}_{i}^T \bs{\mu} \mright)^2\\
& = \sum\limits_{i = p}^{n}\lambda_{p}^{-1} \lambda_{i} \mleft( \mb{v}_{i}^T
\bs{\mu} \mright)^2 - \sum\limits_{i = 1}^{n}\mleft( \mb{v}_{i}^T
\bs{\mu} \mright)^2 - \mleft( \mb{v}_{p}^T \bs{\mu} \mright)^2 - \nu^2\\
& = \sum\limits_{i = p + 1}^{n}\lambda_{p}^{-1} \lambda_{i} \mleft(
\mb{v}_{i}^T \bs{\mu} \mright)^2 - \sum\limits_{i = 1}^{p}\mleft(
\mb{v}_{i}^T \bs{\mu} \mright)^2 - \sum\limits_{i = p + 1}^{n}\mleft(
\mb{v}_{i}^T \bs{\mu} \mright)^2 - \nu^2\\
& = \sum\limits_{i = p + 1}^{n}\mleft( \lambda_{p}^{-1} \lambda_{i} - 1
\mright) \mleft( \mb{v}_{i}^T \bs{\mu} \mright)^2 - \sum\limits_{i =
1}^{p}\mleft( \mb{v}_{i}^T \bs{\mu} \mright)^2 - \nu^2\\
& < 0 \mbox{\ for $\bs{\mu} \ne \mb{0}_n$ or $\nu \ne 0$}
\end{align}
since all coefficients of the quadratic terms are negative, also in the
first term since $\lambda_{p} > \lambda_{i}$ for $i > p$; the quadratic
terms are independent of each other. We observe that the coefficients
coincide with the eigenvalue spectrum obtained from the Jacobian.

We continue with the \textbf{second case} $q > p$
\begin{align}
&\phantom{{}={}} \sum\limits_{i = 1}^{n}\mleft( \lambda_{q}^{-1} \lambda_{i} - 1 \mright)
\mleft( \mb{v}_{i}^T \bs{\mu} \mright)^2 - \mleft( \mb{v}_{q}^T \bs{\mu}
\mright)^2 - \nu^2
\\&\phantom{{}={}}
\nonumber
{} \! - \sum\limits_{i = 1}^{p - 1}\mleft( \mleft[ \lambda_{i} -
\mleft\{ \lambda_{q} + \nu \mright\} \mright]^{-1} + \lambda_{q}^{-1}
\mright) \mleft( \lambda_{i} - \lambda_{q} \mright) \mleft( \mb{v}_{i}^T
\bs{\mu} \mright)^2
\\&\phantom{{}={}}
\nonumber
{} \! + \sum\limits_{i = 1}^{p - 1}\mleft( \mleft[ \lambda_{i} -
\mleft\{ \lambda_{q} + \nu \mright\} \mright]^{-1} + \lambda_{q}^{-1}
\mright) \nu \bs{\mu}^T \mb{v}_{i} \delta_{i,q} \nonumber \\
& = \sum\limits_{i = 1}^{n}\mleft( \lambda_{q}^{-1} \lambda_{i} - 1 \mright)
\mleft( \mb{v}_{i}^T \bs{\mu} \mright)^2 - \mleft( \mb{v}_{q}^T \bs{\mu}
\mright)^2 - \nu^2
\\&\phantom{{}={}}
\nonumber
{} \! - \sum\limits_{i = 1}^{p - 1}\mleft( \mleft[ \lambda_{i} -
\mleft\{ \lambda_{q} + \nu \mright\} \mright]^{-1} + \lambda_{q}^{-1}
\mright) \mleft( \lambda_{i} - \lambda_{q} \mright) \mleft( \mb{v}_{i}^T
\bs{\mu} \mright)^2\\
& \approx \sum\limits_{i = 1}^{n}\mleft( \lambda_{q}^{-1} \lambda_{i} - 1 \mright)
\mleft( \mb{v}_{i}^T \bs{\mu} \mright)^2 - \mleft( \mb{v}_{q}^T \bs{\mu}
\mright)^2 - \nu^2
\\&\phantom{{}={}}
\nonumber
{} \! - \sum\limits_{i = 1}^{p - 1}\mleft( \mleft[ \lambda_{i} -
\lambda_{q} \mright]^{-1} + \lambda_{q}^{-1} \mright) \mleft(
\lambda_{i} - \lambda_{q} \mright) \mleft( \mb{v}_{i}^T \bs{\mu}
\mright)^2\\
& = \sum\limits_{i = 1}^{n}\mleft( \lambda_{q}^{-1} \lambda_{i} - 1 \mright)
\mleft( \mb{v}_{i}^T \bs{\mu} \mright)^2 - \mleft( \mb{v}_{q}^T \bs{\mu}
\mright)^2 - \nu^2 - \sum\limits_{i = 1}^{p - 1}\lambda_{q}^{-1}
\lambda_{i} \mleft( \mb{v}_{i}^T \bs{\mu} \mright)^2\\
& = \sum\limits_{i = 1}^{n}\lambda_{q}^{-1} \lambda_{i} \mleft( \mb{v}_{i}^T
\bs{\mu} \mright)^2 - \sum\limits_{i = 1}^{n}\mleft( \mb{v}_{i}^T
\bs{\mu} \mright)^2 - \mleft( \mb{v}_{q}^T \bs{\mu} \mright)^2 - \nu^2 -
\sum\limits_{i = 1}^{p - 1}\lambda_{q}^{-1} \lambda_{i} \mleft(
\mb{v}_{i}^T \bs{\mu} \mright)^2\\
& = \sum\limits_{i = p}^{n}\lambda_{q}^{-1} \lambda_{i} \mleft( \mb{v}_{i}^T
\bs{\mu} \mright)^2 - \sum\limits_{i = 1}^{p - 1}\mleft( \mb{v}_{i}^T
\bs{\mu} \mright)^2 - \sum\limits_{i = p}^{n}\mleft( \mb{v}_{i}^T
\bs{\mu} \mright)^2 - \mleft( \mb{v}_{q}^T \bs{\mu} \mright)^2 - \nu^2\\
& = \sum\limits_{i = p}^{n}\mleft( \lambda_{q}^{-1} \lambda_{i} - 1 \mright)
\mleft( \mb{v}_{i}^T \bs{\mu} \mright)^2 - \sum\limits_{i = 1}^{p -
1}\mleft( \mb{v}_{i}^T \bs{\mu} \mright)^2 - \mleft( \mb{v}_{q}^T
\bs{\mu} \mright)^2 - \nu^2\\
& = \sum\limits_{i = p}^{q - 1}\mleft( \lambda_{q}^{-1} \lambda_{i} - 1
\mright) \mleft( \mb{v}_{i}^T \bs{\mu} \mright)^2 + \mleft(
\lambda_{q}^{-1} \lambda_{q} - 1 \mright) \mleft( \mb{v}_{q}^T \bs{\mu}
\mright)^2 + \sum\limits_{i = q + 1}^{n}\mleft( \lambda_{q}^{-1}
\lambda_{i} - 1 \mright) \mleft( \mb{v}_{i}^T \bs{\mu} \mright)^2
\\&\phantom{{}={}}
\nonumber
{} \! - \sum\limits_{i = 1}^{p - 1}\mleft( \mb{v}_{i}^T \bs{\mu}
\mright)^2 - \mleft( \mb{v}_{q}^T \bs{\mu} \mright)^2 - \nu^2\\
& = \sum\limits_{i = p}^{q - 1}\mleft( \lambda_{q}^{-1} \lambda_{i} - 1
\mright) \mleft( \mb{v}_{i}^T \bs{\mu} \mright)^2 + \sum\limits_{i = q +
1}^{n}\mleft( \lambda_{q}^{-1} \lambda_{i} - 1 \mright) \mleft(
\mb{v}_{i}^T \bs{\mu} \mright)^2 - \sum\limits_{i = 1}^{p - 1}\mleft(
\mb{v}_{i}^T \bs{\mu} \mright)^2 - \mleft( \mb{v}_{q}^T \bs{\mu}
\mright)^2 - \nu^2.
\end{align}
In the first term, the coefficients of the squared expression are
positive since $\lambda_{q} < \lambda_{i}$ for $i < q$. Since $q > p$,
there is at least one such term, so we have a saddle point. As before,
the quadratic terms are independent of each other. Again we observe the
correspondence between the coefficients and the eigenvalues of the
Jacobian.

We now analyze the critical \textbf{third case} $q < p$ for which the
analysis via the Jacobian failed:
\begin{align}
&\phantom{{}={}} \sum\limits_{i = 1}^{n}\mleft( \lambda_{q}^{-1} \lambda_{i} - 1 \mright)
\mleft( \mb{v}_{i}^T \bs{\mu} \mright)^2 - \mleft( \mb{v}_{q}^T \bs{\mu}
\mright)^2 - \nu^2
\\&\phantom{{}={}}
\nonumber
{} \! - \sum\limits_{i = 1}^{p - 1}\mleft( \mleft[ \lambda_{i} -
\mleft\{ \lambda_{q} + \nu \mright\} \mright]^{-1} + \lambda_{q}^{-1}
\mright) \mleft( \lambda_{i} - \lambda_{q} \mright) \mleft( \mb{v}_{i}^T
\bs{\mu} \mright)^2
\\&\phantom{{}={}}
\nonumber
{} \! + \sum\limits_{i = 1}^{p - 1}\mleft( \mleft[ \lambda_{i} -
\mleft\{ \lambda_{q} + \nu \mright\} \mright]^{-1} + \lambda_{q}^{-1}
\mright) \nu \bs{\mu}^T \mb{v}_{i} \delta_{i,q} \nonumber \\
& = \sum\limits_{i = 1}^{q - 1}\mleft( \lambda_{q}^{-1} \lambda_{i} - 1
\mright) \mleft( \mb{v}_{i}^T \bs{\mu} \mright)^2 + \mleft(
\lambda_{q}^{-1} \lambda_{q} - 1 \mright) \mleft( \mb{v}_{q}^T \bs{\mu}
\mright)^2 + \sum\limits_{i = q + 1}^{n}\mleft( \lambda_{q}^{-1}
\lambda_{i} - 1 \mright) \mleft( \mb{v}_{i}^T \bs{\mu} \mright)^2
\\&\phantom{{}={}}
\nonumber
{} \! - \mleft( \mb{v}_{q}^T \bs{\mu} \mright)^2 - \nu^2
\\&\phantom{{}={}}
\nonumber
{} \! - \sum\limits_{i = 1}^{q - 1}\mleft( \mleft[ \lambda_{i} -
\mleft\{ \lambda_{q} + \nu \mright\} \mright]^{-1} + \lambda_{q}^{-1}
\mright) \mleft( \lambda_{i} - \lambda_{q} \mright) \mleft( \mb{v}_{i}^T
\bs{\mu} \mright)^2
\\&\phantom{{}={}}
\nonumber
{} \! - \mleft( \mleft[ \lambda_{q} - \mleft\{ \lambda_{q} + \nu
\mright\} \mright]^{-1} + \lambda_{q}^{-1} \mright) \mleft( \lambda_{q}
- \lambda_{q} \mright) \mleft( \mb{v}_{q}^T \bs{\mu} \mright)^2
\\&\phantom{{}={}}
\nonumber
{} \! - \sum\limits_{i = q + 1}^{p - 1}\mleft( \mleft[ \lambda_{i} -
\mleft\{ \lambda_{q} + \nu \mright\} \mright]^{-1} + \lambda_{q}^{-1}
\mright) \mleft( \lambda_{i} - \lambda_{q} \mright) \mleft( \mb{v}_{i}^T
\bs{\mu} \mright)^2
\\&\phantom{{}={}}
\nonumber
{} \! + \mleft( \mleft[ \lambda_{q} - \mleft\{ \lambda_{q} + \nu
\mright\} \mright]^{-1} + \lambda_{q}^{-1} \mright) \nu \bs{\mu}^T
\mb{v}_{q}\\
& = \sum\limits_{i = 1}^{q - 1}\mleft( \lambda_{q}^{-1} \lambda_{i} - 1
\mright) \mleft( \mb{v}_{i}^T \bs{\mu} \mright)^2 + \sum\limits_{i = q +
1}^{n}\mleft( \lambda_{q}^{-1} \lambda_{i} - 1 \mright) \mleft(
\mb{v}_{i}^T \bs{\mu} \mright)^2
\\&\phantom{{}={}}
\nonumber
{} \! - \mleft( \mb{v}_{q}^T \bs{\mu} \mright)^2 - \nu^2
\\&\phantom{{}={}}
\nonumber
{} \! - \sum\limits_{i = 1}^{q - 1}\mleft( \mleft[ \lambda_{i} -
\mleft\{ \lambda_{q} + \nu \mright\} \mright]^{-1} + \lambda_{q}^{-1}
\mright) \mleft( \lambda_{i} - \lambda_{q} \mright) \mleft( \mb{v}_{i}^T
\bs{\mu} \mright)^2
\\&\phantom{{}={}}
\nonumber
{} \! - \sum\limits_{i = q + 1}^{p - 1}\mleft( \mleft[ \lambda_{i} -
\mleft\{ \lambda_{q} + \nu \mright\} \mright]^{-1} + \lambda_{q}^{-1}
\mright) \mleft( \lambda_{i} - \lambda_{q} \mright) \mleft( \mb{v}_{i}^T
\bs{\mu} \mright)^2
\\&\phantom{{}={}}
\nonumber
{} \! + \mleft( \mleft[ \lambda_{q} - \mleft\{ \lambda_{q} + \nu
\mright\} \mright]^{-1} + \lambda_{q}^{-1} \mright) \nu \bs{\mu}^T
\mb{v}_{q}\\
& \approx \sum\limits_{i = 1}^{q - 1}\mleft( \lambda_{q}^{-1} \lambda_{i} - 1
\mright) \mleft( \mb{v}_{i}^T \bs{\mu} \mright)^2 + \sum\limits_{i = q +
1}^{n}\mleft( \lambda_{q}^{-1} \lambda_{i} - 1 \mright) \mleft(
\mb{v}_{i}^T \bs{\mu} \mright)^2
\\&\phantom{{}={}}
\nonumber
{} \! - \mleft( \mb{v}_{q}^T \bs{\mu} \mright)^2 - \nu^2
\\&\phantom{{}={}}
\nonumber
{} \! - \sum\limits_{i = 1}^{q - 1}\lambda_{q}^{-1} \lambda_{i} \mleft(
\mb{v}_{i}^T \bs{\mu} \mright)^2
\\&\phantom{{}={}}
\nonumber
{} \! - \sum\limits_{i = q + 1}^{p - 1}\lambda_{q}^{-1} \lambda_{i}
\mleft( \mb{v}_{i}^T \bs{\mu} \mright)^2
\\&\phantom{{}={}}
\nonumber
{} \! - \bs{\mu}^T \mb{v}_{q}\\
& = \mleft( - \sum\limits_{i = 1}^{q - 1}\mleft[ \mb{v}_{i}^T \bs{\mu}
\mright]^2 \mright) + \sum\limits_{i = q + 1}^{p - 1}\mleft(
\lambda_{q}^{-1} \lambda_{i} - 1 \mright) \mleft( \mb{v}_{i}^T \bs{\mu}
\mright)^2 + \sum\limits_{i = p}^{n}\mleft( \lambda_{q}^{-1} \lambda_{i}
- 1 \mright) \mleft( \mb{v}_{i}^T \bs{\mu} \mright)^2
\\&\phantom{{}={}}
\nonumber
{} \! - \mleft( \mb{v}_{q}^T \bs{\mu} \mright)^2 - \nu^2 -
\sum\limits_{i = q + 1}^{p - 1}\lambda_{q}^{-1} \lambda_{i} \mleft(
\mb{v}_{i}^T \bs{\mu} \mright)^2 - \bs{\mu}^T \mb{v}_{q}\\
& = \mleft( - \sum\limits_{i = 1}^{q}\mleft[ \mb{v}_{i}^T \bs{\mu}
\mright]^2 \mright) - \sum\limits_{i = q + 1}^{p - 1}\mleft(
\mb{v}_{i}^T \bs{\mu} \mright)^2 + \sum\limits_{i = p}^{n}\mleft(
\lambda_{q}^{-1} \lambda_{i} - 1 \mright) \mleft( \mb{v}_{i}^T \bs{\mu}
\mright)^2 - \nu^2 - \bs{\mu}^T \mb{v}_{q}\\
& = \mleft( - \sum\limits_{i = 1}^{p - 1}\mleft[ \mb{v}_{i}^T \bs{\mu}
\mright]^2 \mright) + \sum\limits_{i = p}^{n}\mleft( \lambda_{q}^{-1}
\lambda_{i} - 1 \mright) \mleft( \mb{v}_{i}^T \bs{\mu} \mright)^2 -
\nu^2 - \bs{\mu}^T \mb{v}_{q}.
\end{align}
We see (incomplete) correspondences between the coefficients of the
first three terms (which are all negative) and the eigenvalues of the
Jacobian determined above. However, we now have the additional term $-
\bs{\mu}^T \mb{v}_{q} $. This term can become positive, e.g. if
$\bs{\mu}$ points into the direction $- \mb{v}_{q}$. If we insert
$\bs{\mu} = - \varepsilon \mb{v}_{q} $ with small $\varepsilon$, only
the first and last summand survive, and the term with $\nu$ has to be
preserved since it can't be zero in this case (as this would result in
undefined terms in the differential equation):
\begin{align}
&\phantom{{}={}} \mleft( - \sum\limits_{i = 1}^{p - 1}\mleft[ \mb{v}_{i}^T \bs{\mu}
\mright]^2 \mright) + \sum\limits_{i = p}^{n}\mleft( \lambda_{q}^{-1}
\lambda_{i} - 1 \mright) \mleft( \mb{v}_{i}^T \bs{\mu} \mright)^2 -
\nu^2 - \bs{\mu}^T \mb{v}_{q} \nonumber \\
& = \mleft( - \varepsilon^2 \mright) + \varepsilon - \nu^2\\
& \approx \varepsilon - \nu^2
\end{align}
which becomes positive for $\varepsilon > \nu^2$. Therefore the critical
points under consideration are saddle points.

Over all three cases, this shows that the system converges to the
desired fixed point $p$.

\subsection{Learning Rule System}

%
To derive the learning rule system, we continue from
\eqref{eq_newton_p_1}, replace $\mb{w} = \mb{w}_{p}$ and $l = l_{p}$,
replace true eigenvectors by estimates $\mb{v}_{i} \approx \mb{w}_{i}$,
and replace true eigenvalues by estimates $\lambda_{i} \approx l_{i}$:
\begin{align}
\label{eq_wdot_arb}
\dot{\mb{w}}_{p}
&=
l_{p}^{-1} \mleft( \mb{C} \mb{w}_{p} - \mleft[ \mb{w}_{p}^T \mb{C}
\mb{w}_{p} \mright] \mb{w}_{p} \mright) + \frac{1}{2}\, \mb{w}_{p}
\mleft( \mb{w}_{p}^T \mb{w}_{p} - 1 \mright)
\\&\phantom{{}={}}
\nonumber
{} \! - \mleft( \sum\limits_{i = 1}^{p - 1}\mleft[ \mleft\{ l_{i} -
l_{p} \mright\}^{-1} + l_{p}^{-1} \mright] \mb{w}_{i} \mb{w}_{i}^T
\mright) \mleft( \mb{C} \mb{w}_{p} - l_{p} \mb{w}_{p} \mright).
\end{align}
For the eigenvalue estimates, this leads to
\begin{align}
\label{eq_ldot_arb}
\dot{l}_{p}
&=
\mb{w}_{p}^T \mb{C} \mb{w}_{p} - l_{p} \mb{w}_{p}^T \mb{w}_{p}.
\end{align}
We see no way how \eqref{eq_wdot_arb} and \eqref{eq_ldot_arb} could be
turned into matrix form.

\subsection{Simulations}

%
For the learning rule system \eqref{eq_wdot_arb} \eqref{eq_ldot_arb} we
performed simulations with artificial data using Octave. The problem
dimension was chosen as $n = 10$, the subspace dimension as $m = 5$. A
covariance matrix was determined from a randomly generated orthogonal
matrix of true eigenvectors and a log-linear set of true eigenvalues
($\exp(-i)$ with index $i$).

\textbf{Stability experiment:} We use the learning rule for the desired
fixed point $p$. The system state is computed by adding a small
perturbation to fixed point $q$. We determine the scalar product between
the permutation and the direction vector of the learning rule system. By
repeating this computation $100.000$ times for different perturbations,
we observe instability for $p \ne q$ (scalar product is sometimes
positive) and stability for $p = q$ (scalar product is never positive),
confirming the analytical stability results.

\textbf{Numerical solution:} We solve the learning rule system
numerically by applying Euler's method for $100.000$ steps with a step
width $\gamma = 10^{-3}$. The previous $p - 1$ eigenvector and
eigenvalue estimates are set to the true values. We randomly initialize
the $p$-th eigenvector estimate (normalized to unit length). The
behavior differs depending on the initial $p$-th eigenvalue estimate.
With small initial eigenvalue estimates, the system typically converges
to the desired fixed point. If the initial eigenvalue estimates are too
small, the system sometimes diverges and delivers ``not a number'' (NaN)
solutions. With large initial eigenvalue estimates, the system sometimes
converges to a zero eigenvector estimate with an apparently arbitrary
eigenvalue estimate. As pointed out in section
\ref{sec_newton_arbitrary}), there is a set of additional fixed points
introduced by applying Newton's method, and the simulations show that
these fixed points appear to be semi-stable. We observed that a
normalization of the $p$-th eigenvector estimates to unit length in each
step guarantees convergence to the desired fixed point. We assume that
the instabilities are caused by large values obtained in the inverse
eigenvalue terms. If we start with random eigenvalue estimates in the
vicinity of the eigenvalue in the desired fixed point, the learning rule
system converges to the desired fixed point.

In summary, the learning rule system works as expected in the vicinity
of the desired fixed point, but has stability problems when started in a
larger distance.

\subsection{Relation to Deflation}

%
We see a structural similarity of the learning rule systems derived for
the principal eigenpair (with deflation) and the learning rule system
derived for the arbitrary eigenpair. If we compare the deflation-related
terms in $\dot{\mb{w}}_{p}$ from \eqref{eq_wdot_defl} (estimate of
principal eigenpair and deflation)
\begin{align}
\mleft( - \sum\limits_{i = 1}^{p - 1}l_{i} l_{p}^{-1} \mb{w}_{i}
\mb{w}_{i}^T \mb{w}_{p} \mright) + \sum\limits_{i = 1}^{p - 1}l_{i}
l_{p}^{-1} \mleft( \mb{w}_{i}^T \mb{w}_{p} \mright)^2 \mb{w}_{p}
\end{align}
with the related terms in $\dot{\mb{w}}_{p}$ from \eqref{eq_wdot_arb}
(arbitrary eigenpair)
\begin{align}
\mleft( - \sum\limits_{i = 1}^{p - 1}\mleft[ \mleft\{ l_{i} - l_{p}
\mright\}^{-1} + l_{p}^{-1} \mright] \mb{w}_{i} \mb{w}_{i}^T \mb{C}
\mb{w}_{p} \mright) + \sum\limits_{i = 1}^{p - 1}\mleft( \mleft[ l_{i} -
l_{p} \mright]^{-1} l_{p} + 1 \mright) \mb{w}_{i} \mb{w}_{i}^T
\mb{w}_{p},
\end{align}
we also see similarities, but the mechanisms are clearly different with
not obvious mapping between them. There is a further difference in that
the deflation-related terms for $\dot{l}_{p}$ in \eqref{eq_ldot_defl}
are absent from \eqref{eq_ldot_arb}.

It is presently unclear whether there are differences with respect to
convergence speed or stability for larger learning rates between the two
versions.

\section{Conclusions and Future Work}\label{sec_conclusion}

%
This work describes a Lagrange-Newton framework for the derivation of
coupled learning rules which we feel is superior to our earlier two
attempts at devising suitable frameworks. For the example of PCA we
could demonstrate for two different cases that learning rule systems
with the desired favorable convergence properties can be derived from
this framework. The first case --- a derivation for the {\em principal}
eigenpair --- leads to the same result as the derivations from our
earlier frameworks. The second case --- a derivation for an {\em
arbitrary eigenpair} --- leads to novel learning rules with an
interesting alternative to deflation.

A stability analysis for exact Hessians already indicates that --- at
least for the PCA case --- the framework leads to learning rules which
only have stable fixed points at the desired location, i.e. at those
fixed points for which the Hessian was determined; all other locations
are saddle points. However, exact Hessians are unknown when learning
rules are applied since they can only be determined if the eigenpair
that needs to be estimated is already known. The usefulness of our
framework is furthermore confirmed by the stability analysis for the two
cases mentioned above which confirms that both learning rules converge
to the desired fixed points. However, earlier attempts revealed that
undesired fixed points can appear depending on which approximations are
used in the derivation. Since the effect of approximations on the
stability is difficult to predict, a stability analysis is always
required.

Simulations for the second case (the first one has been studied before)
are in accordance with the results of the stability analysis. However,
depending on the initial values, the learning rule system may converge
to an additional set of fixed points which is introduced by the
multiplication with the inverse Hessian, since the Hessian is undefined
at these points. It is well known that Newton's method only works in the
vicinity of the fixed point but can diverge in larger distance. Future
work will therefore explore whether the learning rule needs to be
modified in larger distance from the fixed point, e.g. by a
Levenberg-Marquardt modification \citep[see e.g.][]{nn_Chong01}.

We encountered a pitfall in the stability analysis of the second case:
for one sub-case, the eigenvalues of the Jacobian indicated a stable
undesired fixed point which was in conflict with the simulation results.
The reason was an undefined term which, however, later disappeared from
the derivation. We therefore had to repeat the stability analysis with a
perturbation approach which correctly indicates a saddle point.

In future work we will attempt to apply the Lagrange-Newton framework to
symmetric rules as we have studied before
\citep[][]{own_Moeller20a,own_Moeller20b}.
%
%

%

\appendix

\section{Lemmata}

\subsection{Lagrange Multiplier Method}

%
\textbf{Dimensions:}
\begin{description}
\setlength{\itemsep}{0pt}
\item[] $n$: dimensionality of optimization problem, $3 \leq n$
\item[] $m$: number of equality constraints, $1 \leq m$
\end{description}
\textbf{Matrices:}
\begin{description}
\setlength{\itemsep}{0pt}
\item[] $\mb{x}$: arbitrary, $n \times 1$
\item[] $f\mleft( \mb{x} \mright)$: objective function, scalar, $1 \times 1$
\item[] $\mb{g}\mleft( \mb{x} \mright)$: vector of equality constraints, arbitrary, $m \times 1$
\item[] $g_{j}\mleft( \mb{x} \mright)$: single equality constraint, scalar, $1 \times 1$, vector element of $\mb{g}$
\item[] $\bs{\lambda}$: vector of Lagrange multipliers, arbitrary, $m \times 1$
\item[] $\lambda_{j}\mleft( \mb{x} \mright)$: single Lagrange multiplier, scalar, $1 \times 1$, vector element of $\bs{\lambda}$
\item[] $L$: Lagrangian, scalar, $1 \times 1$
\item[] $\bar{\mb{H}}$: bordered Hessian, symmetric, $(n + m) \times (n + m)$
\end{description}

\subsubsection{Bordered Hessian}\label{sec_bordered_hessian}

%
In the following we explore whether the extended objective function
(over an extended variable vector containing the original variables and
the Lagrange multipliers) has saddle points at all fixed points. We
currently can't provide a general statement on the eigenvalues, but only
analyze the special case of PCA.

We start by deriving the general bordered Hessian. Given an objective
function $f\mleft( \mb{x} \mright)$, equality constraints $\mb{g}\mleft(
\mb{x} \mright)$, and Lagrange multipliers $\bs{\lambda}$, the
Lagrangian of the constrained optimization problem is defined as
\begin{align}
L\mleft( \mb{x},\bs{\lambda} \mright)
&=
f\mleft( \mb{x} \mright) + \bs{\lambda}^T \mb{g}\mleft( \mb{x} \mright).
\end{align}
The first-order derivatives (gradient of the Lagrangian) are
\begin{align}
\frac{\partial}{\partial \mb{x}}L\mleft( \mb{x},\bs{\lambda} \mright)
&=
\frac{\partial}{\partial \mb{x}}f\mleft( \mb{x} \mright) +
\bs{\lambda}^T \mleft( \frac{\partial}{\partial \mb{x}}\mb{g}\mleft(
\mb{x} \mright) \mright)\\
\frac{\partial}{\partial \bs{\lambda}}L\mleft( \mb{x},\bs{\lambda}
\mright)
&=
\mleft( \mb{g}\mleft( \mb{x} \mright) \mright)^T.
\end{align}
To form the bordered Hessian
\begin{align}
\bar{\mb{H}}
&=
\begin{pmatrix}\frac{\partial}{\partial \mb{x}}\mleft(
\frac{\partial}{\partial \mb{x}}L\mleft( \mb{x},\bs{\lambda} \mright)
\mright)^T & \frac{\partial}{\partial \bs{\lambda}}\mleft(
\frac{\partial}{\partial \mb{x}}L\mleft( \mb{x},\bs{\lambda} \mright)
\mright)^T\\\frac{\partial}{\partial \mb{x}}\mleft(
\frac{\partial}{\partial \bs{\lambda}}L\mleft( \mb{x},\bs{\lambda}
\mright) \mright)^T & \frac{\partial}{\partial \bs{\lambda}}\mleft(
\frac{\partial}{\partial \bs{\lambda}}L\mleft( \mb{x},\bs{\lambda}
\mright) \mright)^T\end{pmatrix}
\end{align}
we need to determine the second-order derivatives in the four blocks. We
start with the upper-left block:
\begin{align}
&\phantom{{}={}} \frac{\partial}{\partial \mb{x}}\mleft( \frac{\partial}{\partial
\mb{x}}L\mleft( \mb{x},\bs{\lambda} \mright) \mright)^T \nonumber \\
& = \frac{\partial}{\partial \mb{x}}\mleft( \frac{\partial}{\partial
\mb{x}}f\mleft( \mb{x} \mright) \mright)^T + \frac{\partial}{\partial
\mb{x}}\mleft( \bs{\lambda}^T \mleft[ \frac{\partial}{\partial
\mb{x}}\mb{g}\mleft( \mb{x} \mright) \mright] \mright)^T\\
& = \frac{\partial}{\partial \mb{x}}\mleft( \frac{\partial}{\partial
\mb{x}}f\mleft( \mb{x} \mright) \mright)^T + \frac{\partial}{\partial
\mb{x}}\mleft( \mleft[ \frac{\partial}{\partial \mb{x}}\mb{g}\mleft(
\mb{x} \mright) \mright]^T \bs{\lambda} \mright).
\end{align}
For the second term, we cannot find a closed-form matrix expression, but
we can compute the derivative for row $i$ according to
\eqref{eq_ddx_BxTa}
\begin{align}
&\phantom{{}={}} \mleft( \frac{\partial}{\partial \mb{x}}\mleft[ \mleft\{
\frac{\partial}{\partial \mb{x}}\mb{g}\mleft( \mb{x} \mright)
\mright\}^T \bs{\lambda} \mright] \mright)_{i,*} \nonumber \\
& = \bs{\lambda}^T \mleft( \frac{\partial}{\partial \mb{x}}\mleft[
\frac{\partial}{\partial x_{i}}\mb{g}\mleft( \mb{x} \mright) \mright]
\mright)
\end{align}
which in turn can be expressed in single-element form as
\begin{align}
\frac{\partial}{\partial \mb{x}}\mleft( \mleft[ \frac{\partial}{\partial
\mb{x}}\mb{g}\mleft( \mb{x} \mright) \mright]^T \bs{\lambda} \mright)
&=
\mleft( \sum\limits_{k = 1}^{m}\lambda_{k} \mleft[
\frac{\partial}{\partial x_{j}}\mleft\{ \frac{\partial}{\partial
x_{i}}g_{k}\mleft( \mb{x} \mright) \mright\} \mright] \mright)^{n\times
n}_{i,j},
\end{align}
thus we get
\begin{align}
\frac{\partial}{\partial \mb{x}}\mleft( \frac{\partial}{\partial
\mb{x}}L\mleft( \mb{x},\bs{\lambda} \mright) \mright)^T
&=
\frac{\partial}{\partial \mb{x}}\mleft( \frac{\partial}{\partial
\mb{x}}f\mleft( \mb{x} \mright) \mright)^T + \mleft( \sum\limits_{k =
1}^{m}\lambda_{k} \mleft[ \frac{\partial}{\partial x_{j}}\mleft\{
\frac{\partial}{\partial x_{i}}g_{k}\mleft( \mb{x} \mright) \mright\}
\mright] \mright)^{n\times n}_{i,j}.
\end{align}
The other three blocks of the bordered Hessian are easier to determine
(note that a Hessian is always symmetric):
\begin{align}
\frac{\partial}{\partial \bs{\lambda}}\mleft( \frac{\partial}{\partial
\mb{x}}L\mleft( \mb{x},\bs{\lambda} \mright) \mright)^T
&=
\mleft( \frac{\partial}{\partial \mb{x}}\mb{g}\mleft( \mb{x} \mright)
\mright)^T
\end{align}
\begin{align}
\frac{\partial}{\partial \mb{x}}\mleft( \frac{\partial}{\partial
\bs{\lambda}}L\mleft( \mb{x},\bs{\lambda} \mright) \mright)^T
&=
\frac{\partial}{\partial \mb{x}}\mb{g}\mleft( \mb{x} \mright)
\end{align}
\begin{align}
\frac{\partial}{\partial \bs{\lambda}}\mleft( \frac{\partial}{\partial
\bs{\lambda}}L\mleft( \mb{x},\bs{\lambda} \mright) \mright)^T
&=
\mb{0}_{m,m}.
\end{align}
The eigenvalues of the bordered Hessian at a given fixed point determine
the stability of this fixed point.

I'm not aware of a general statement on the eigenvalues of the bordered
Hessian. In the following, we apply the derivation to the PCA objective
function \eqref{eq_J_pca} which we can rewrite in this notation as
\begin{align}
L\mleft( \mb{x},\lambda \mright)
&=
\frac{1}{2}\, \mb{x}^T \mb{C} \mb{x} - \frac{1}{2}\, \lambda \mleft(
\mb{x}^T \mb{x} - 1 \mright)
\end{align}
thus
\begin{align}
f\mleft( \mb{x} \mright)
&=
\frac{1}{2}\, \mb{x}^T \mb{C} \mb{x}\\
g\mleft( \mb{x} \mright)
&=
- \frac{1}{2}\, \mleft( \mb{x}^T \mb{x} - 1 \mright).
\end{align}
The corresponding unconstrained optimization (minimization) problem is
\begin{align}
\hat{f}\mleft( \mb{x} \mright)
&=
- \frac{1}{2}\, \mb{x}^T \mb{C} \mb{x}
\end{align}
and the corresponding Hessian is $- \mb{C}$. Since $\mb{C}$ is a
covariance matrix which is positive semi-definite, all eigenvalues of
the Hessian are negative or zero.

The bordered Hessian of the constrained optimization problem (over the
extended variable vector) is
\begin{align}
\bar{\mb{H}}
&=
\begin{pmatrix}\mb{C} - \lambda \mb{I}_{n} & - \mb{x}\\- \mb{x}^T &
0\end{pmatrix}.
\end{align}
We can proceed as in section \ref{sec_hessian}: At a fixed point $p$, we
get a transformed Hessian
\begin{align}
{\mb{\bar{H}^*}}
&=
\begin{pmatrix}\bs{\Lambda} - \lambda_{p} \mb{I}_{n} & - \mb{e}_{p}\\-
\mb{e}_{p}^T & 0\end{pmatrix}.
\end{align}
Its eigenvalues $s$ are determined from
\begin{align}
&\phantom{{}={}} \opnl{det} \mleft\{ {\mb{\bar{H}^*}} - s \mb{I}_{n + 1} \mright\} \nonumber \\
& = \resizebox{\minof{\width}{\symaeqwidth}}{!}{%
$\begin{vmatrix}\mleft( \lambda_{1} - \lambda_{p} \mright) - s & & & & &
& & \\ & \ddots & & & & & & \\ & & \mleft( \lambda_{p - 1} - \lambda_{p}
\mright) - s & & & & & \\ & & & - s & & & & - 1\\ & & & & \mleft(
\lambda_{p + 1} - \lambda_{p} \mright) - s & & & \\ & & & & & \ddots & &
\\ & & & & & & \mleft( \lambda_{n} - \lambda_{p} \mright) - s & \\ & & &
- 1 & & & & - s\end{vmatrix}_{n + 1}$%
}\\
& = 0
\end{align}
We obtain the eigenvalues $s$ by analyzing under which conditions entire
rows become zero, or under which conditions adding multiples of rows to
other rows produces zero rows, and obtain:
\begin{align}
s_{k}
&=
\lambda_{k} - \lambda_{p} \quad\mbox{for}\;k=1,\ldots,n,\;k \neq p\\
s_{p}
&=
- 1\\
s_{n + 1}
&=
1.
\end{align}
Already $s_{p}$ and $s_{n + 1}$ indicate that all fixed points are
saddle points. Moreover, for $p \neq 1$ and $p \neq n$, we have a mix of
positive and negative values for $s_{k}$ (note that we assume descending
eigenvalues $\lambda_{k}$ in $\bs{\Lambda}$).

Neither following the positive nor following the negative gradient will
converge to any fixed point (if we would have only positive or only
negative eigenvalues, we would get convergence for either the negative
or the positive gradient).

\subsection{Derivatives}\label{app_derivatives}

%
Derivatives presented in the following without giving a derivation were
computed using the tool at \url{www.matrixcalculus.org}. Note that
according to our definition of partial derivatives, the gradient is a
row vector (transposed relative to the output of the tool) and the
derivative of a vector with respect to a vector produces the standard
Jacobian form (same as output of the tool) \citep[overall we use the
same convention as in][p.127]{nn_Deisenroth20}.

Given index $n$: $1 \leq n$ and matrices $\mb{x}$: arbitrary, $n \times
1$; $\mb{A}$: square, $n \times n$; $\mb{S}$: symmetric, $n \times n$,
we have:
\begin{align}
\frac{\partial}{\partial \mb{x}}\mleft( \mleft[ \mb{x}^T \mb{S} \mb{x}
\mright] \mb{x} \mright)
&=
2 \mb{x} \mb{x}^T \mb{S} + \mleft( \mb{x}^T \mb{S} \mb{x} \mright)
\mb{I}_{n}\\
\frac{\partial}{\partial \mb{x}}\mleft( \mleft[ \mb{x}^T \mb{x} \mright]
\mb{x} \mright)
&=
2 \mb{x} \mb{x}^T + \mleft( \mb{x}^T \mb{x} \mright) \mb{I}_{n}
\end{align}
Given indices $n$: $1 \leq n$; $m$: $1 \leq m$; $i$: $1 \leq i \leq n$
and matrices $\mb{x}$: arbitrary, $n \times 1$; $\mb{a}$: constant,
arbitrary, $m \times 1$; $\mb{B}\mleft( \mb{x} \mright)$: function of
$\mb{x}$, arbitrary, $m \times n$; $\mb{b}_{i}\mleft( \mb{x} \mright)$:
function of $\mb{x}$, arbitrary, $m \times 1$, column vector of
$\mb{B}$, we have
\begin{align}
&\phantom{{}={}} \mleft( \frac{\partial}{\partial \mb{x}}\mleft[ \mleft\{ \mb{B}\mleft(
\mb{x} \mright) \mright\}^T \mb{a} \mright] \mright)_{i,*} \nonumber \\
& = \frac{\partial}{\partial \mb{x}}\mleft( \mleft[ \mb{B}\mleft( \mb{x}
\mright) \mright]^T \mb{a} \mright)_{i}\\
& = \frac{\partial}{\partial \mb{x}}\mleft( \mleft[ \mb{b}_{i}\mleft( \mb{x}
\mright) \mright]^T \mb{a} \mright)\\
&\label{eq_ddx_BxTa}
 = \mb{a}^T \mleft( \frac{\partial}{\partial \mb{x}}\mb{b}_{i}\mleft(
\mb{x} \mright) \mright).
\end{align}

\subsection{Triangular Matrices}

\begin{lemma}\label{lemma_sut_basic}
Given indices $m$: $2 \leq m \leq n$; $p$: $2 \leq p \leq m$ and matrix
$\mb{A}$: square, $m \times m$, we have for column $p$ of a strictly
upper triangular matrix:
\begin{align}
\mleft( \opnl{sut} \mleft\{ \mb{A} \mright\} \mright)_{p}
&=
\begin{pmatrix}\mb{I}_{p - 1} & \mb{0}_{p - 1,m - p + 1}\\\mb{0}_{m - p
+ 1,p - 1} & \mb{0}_{m - p + 1,m - p + 1}\end{pmatrix} \mleft( \mb{A}
\mright)_{p}
\end{align}
and for the special case $p = 1$
\begin{align}
\mleft( \opnl{sut} \mleft\{ \mb{A} \mright\} \mright)_{1}
&=
\mb{0}_{m}
\end{align}
\end{lemma}
\begin{lemma}\label{lemma_sut_triple}
Given indices $m$: $2 \leq m \leq n$; $p$: $2 \leq p \leq m$ and
matrices $\mb{U}$: arbitrary, $n \times m$; $\mb{V}$: arbitrary, $n
\times m$; $\mb{W}$: arbitrary, $n \times m$, we have with Lemma
\ref{lemma_sut_basic} for column $p$ of the product of a matrix with a
strictly upper triangular matrix of a matrix product:
\begin{align}
&\phantom{{}={}} \mleft( \mb{U} \opnl{sut} \mleft\{ \mb{V}^T \mb{W} \mright\}
\mright)_{p} \nonumber \\
& = \mb{U} \mleft( \opnl{sut} \mleft\{ \mb{V}^T \mb{W} \mright\}
\mright)_{p}\\
& = \mb{U} \mleft( \begin{pmatrix}\mb{I}_{p - 1} & \mb{0}_{p - 1,m - p +
1}\\\mb{0}_{m - p + 1,p - 1} & \mb{0}_{m - p + 1,m - p + 1}\end{pmatrix}
\mleft( \mb{V}^T \mb{W} \mright)_{p} \mright)\\
& = \mb{U} \mleft( \begin{pmatrix}\mb{I}_{p - 1} & \mb{0}_{p - 1,m - p +
1}\\\mb{0}_{m - p + 1,p - 1} & \mb{0}_{m - p + 1,m - p + 1}\end{pmatrix}
\mleft[ \mb{V}^T \mb{w}_{p} \mright] \mright)\\
& = \mleft( \mb{U} \begin{pmatrix}\mb{I}_{p - 1} & \mb{0}_{p - 1,m - p +
1}\\\mb{0}_{m - p + 1,p - 1} & \mb{0}_{m - p + 1,m - p + 1}\end{pmatrix}
\mright) \mleft( \mb{V}^T \mb{w}_{p} \mright)\\
& = \sum\limits_{i = 1}^{m}\mleft( \mb{U} \begin{pmatrix}\mb{I}_{p - 1} &
\mb{0}_{p - 1,m - p + 1}\\\mb{0}_{m - p + 1,p - 1} & \mb{0}_{m - p + 1,m
- p + 1}\end{pmatrix} \mright)_{i} \mleft( \mb{V}^T \mb{w}_{p}
\mright)_{i}\\
& = \sum\limits_{i = 1}^{m}\mleft( \mb{U} \begin{pmatrix}\mb{I}_{p - 1} &
\mb{0}_{p - 1,m - p + 1}\\\mb{0}_{m - p + 1,p - 1} & \mb{0}_{m - p + 1,m
- p + 1}\end{pmatrix} \mright)_{i} \mleft( \mb{v}_{i}^T \mb{w}_{p}
\mright)\\
& = \sum\limits_{i = 1}^{p - 1}\mb{u}_{i} \mleft( \mb{v}_{i}^T \mb{w}_{p}
\mright)\\
& = \sum\limits_{i = 1}^{p - 1}\mb{u}_{i} \mb{v}_{i}^T \mb{w}_{p}
\end{align}
and for the special case ($p = 1$), the left side is $\mb{0}_{m}$ and
the right side is the empty sum and therefore $\mb{0}_{m}$ as well.
\end{lemma}
\begin{lemma}\label{lemma_sut_diag}
Given indices $m$: $2 \leq m \leq n$; $p$: $2 \leq p \leq m$ and
matrices $\mb{A}$: square, $m \times m$; $\mb{D}$: diagonal, $m \times
m$, we have with Lemma \ref{lemma_sut_basic}:
\begin{align}
\opnl{sut} \mleft\{ \mb{A} \mb{D} \mright\}
&=
\opnl{sut} \mleft\{ \mb{A} \mright\} \mb{D}
\end{align}
since
\begin{align}
&\phantom{{}={}} \mleft( \opnl{sut} \mleft\{ \mb{A} \mb{D} \mright\} \mright)_{p} \nonumber \\
& = \begin{pmatrix}\mb{I}_{p - 1} & \mb{0}_{p - 1,m - p + 1}\\\mb{0}_{m - p
+ 1,p - 1} & \mb{0}_{m - p + 1,m - p + 1}\end{pmatrix} \mleft( \mb{A}
\mb{D} \mright)_{p}\\
& = \begin{pmatrix}\mb{I}_{p - 1} & \mb{0}_{p - 1,m - p + 1}\\\mb{0}_{m - p
+ 1,p - 1} & \mb{0}_{m - p + 1,m - p + 1}\end{pmatrix} \mb{A} \mleft(
\mb{D} \mright)_{p}\\
& = \mleft( \begin{pmatrix}\mb{I}_{p - 1} & \mb{0}_{p - 1,m - p +
1}\\\mb{0}_{m - p + 1,p - 1} & \mb{0}_{m - p + 1,m - p + 1}\end{pmatrix}
\mb{A} \mright) \mleft( \mb{D} \mright)_{p}\\
& = \sum\limits_{i = 1}^{m}\mleft( \begin{pmatrix}\mb{I}_{p - 1} & \mb{0}_{p
- 1,m - p + 1}\\\mb{0}_{m - p + 1,p - 1} & \mb{0}_{m - p + 1,m - p +
1}\end{pmatrix} \mb{A} \mright)_{i} \mleft( \mleft( \mb{D} \mright)_{p}
\mright)_{i}\\
& = \sum\limits_{i = 1}^{m}\mleft( \begin{pmatrix}\mb{I}_{p - 1} & \mb{0}_{p
- 1,m - p + 1}\\\mb{0}_{m - p + 1,p - 1} & \mb{0}_{m - p + 1,m - p +
1}\end{pmatrix} \mleft( \mb{A} \mright)_{i} \mright) \mleft( \mleft(
\mb{D} \mright)_{p} \mright)_{i}\\
& = \sum\limits_{i = 1}^{m}\mleft( \opnl{sut} \mleft\{ \mb{A} \mright\}
\mright)_{i} \mleft( \mleft( \mb{D} \mright)_{p} \mright)_{i}\\
& = \opnl{sut} \mleft\{ \mb{A} \mright\} \mleft( \mb{D} \mright)_{p}\\
& = \mleft( \opnl{sut} \mleft\{ \mb{A} \mright\} \mb{D} \mright)_{p}
\end{align}
and the special case $p = 1$ is obvious.
\end{lemma}

\end{document}